\documentclass[preprint,3p,12pt]{elsarticle}
\usepackage{layout}
\usepackage{amsmath,amssymb,amsfonts}
\usepackage{algorithmic}
\usepackage{graphicx}
\usepackage{xcolor}
\usepackage{textcomp}
\usepackage{booktabs}
\usepackage{caption}
\usepackage{comment}
\usepackage{float}
\usepackage{multirow}
\usepackage{makecell}
\usepackage{pdflscape}
\usepackage{microtype}
\usepackage{url}
\usepackage[acronym]{glossaries}
\usepackage{subcaption}
\usepackage{geometry}
\usepackage{lineno}
\makeglossaries
\newacronym{LLM}{LLM}{Large Language Model}
\newacronym{LLMs}{LLMs}{Large Language Models}
\newacronym{VLMs}{VLMs}{Vision–Language Models}
\newacronym{LMMs}{LMMs}{Large Multimodal Models}
\newacronym{VQA}{VQA}{Visual Question Answering}
\newacronym{DL}{DL}{Deep Learning}
\newacronym{AI}{AI}{Artificial Intelligence}
\newacronym{CDDM}{CDDM}{Crop disease domain multimodal dataset}
\newacronym{LVLMs}{LVLMs}{Large Vision-Language Models}
\newacronym{AUC}{AUC}{Area Under the Curve}
\begin{document}

\begin{frontmatter}



\title{LeafNet: A Large-Scale Dataset and Comprehensive Benchmark for Foundational Vision-Language Understanding of Plant Diseases}


\author[1]{Khang Nguyen Quoc}
\affiliation[1]{organization={School of Electrical Engineering, Korea University},
            city={Seoul},
            postcode={02841},
            country={South Korea}}

\author[2]{Phuong D. Dao\corref{cor1}}
\ead{phuong.dao@austin.utexas.edu}
\affiliation[2]{organization={Department of Integrative Biology, The University of Texas at Austin},
            city={Austin},
            postcode={TX 78712},
            country={USA}}

\author[3]{Luyl-Da Quach\corref{cor1}}
\ead{luyldaquach@gmail.com}
\affiliation[3]{organization={Department of Software Engineering, FPT University},
            city={Cantho city},
            postcode={CT 90000},
            country={Vietnam}}

% Define the corresponding author text (usually required if you use \corref)
\cortext[cor1]{Corresponding author}

%% Abstract
\begin{abstract}
Foundation models and vision-language pre-training have significantly advanced \gls{VLMs}, enabling multimodal processing of visual and linguistic data. However, their application in domain-specific agricultural tasks, such as plant pathology, remains limited due to the lack of large-scale, comprehensive multimodal image–text datasets and benchmarks. To address this gap, we introduce LeafNet, a comprehensive multimodal dataset, and LeafBench, a visual question-answering benchmark developed to systematically evaluate the capabilities of \gls{VLMs} in understanding plant diseases. The dataset comprises 186,000 leaf digital images spanning 97 disease classes, paired with comprehensive expert-annotated metadata, generating 13,950 question-answer pairs spanning six critical agricultural tasks: crop species identification, healthy-diseased classification, disease identification, symptom identification, pathogen classification, and taxonomic name classification. The questions assess various aspects of plant pathology understanding, including visual symptom recognition, taxonomic relationships, and diagnostic reasoning. Benchmarking 12 state-of-the-art \gls{VLMs} on our LeafBench dataset, including GPT-4o and Gemini 2.5 Pro, we reveal substantial disparity in their disease understanding capabilities. Closed-source models achieved up to 72\% accuracy, whereas generic open-source VLMs struggle with near-chance performance. Our study shows performance varies markedly across tasks: binary healthy–diseased classification exceeds 90\% accuracy, while fine-grained pathogen and species identification remains below 65\%. Direct comparison between vision-only models and \gls{VLMs} demonstrates the critical advantage of multimodal architectures: fine-tuned \gls{VLMs} outperform traditional vision models by up to 27.76\% on semantically demanding tasks, with domain-specific models like SCOLD achieving 99.15\% disease identification accuracy and 94.92\% symptom identification accuracy, confirming that integrating linguistic representations significantly enhances diagnostic precision. These findings highlight critical gaps in current \gls{VLMs} for plant pathology applications and underscore the need for LeafBench as a rigorous framework for methodological advancement and progress evaluation toward reliable \gls{AI}-assisted plant disease diagnosis. Codes will be available at \url{https://github.com/EnalisUs/LeafBench}.
\end{abstract}

\begin{graphicalabstract}
\begin{figure}[H]
 \centering
 \includegraphics[width=1\textwidth]{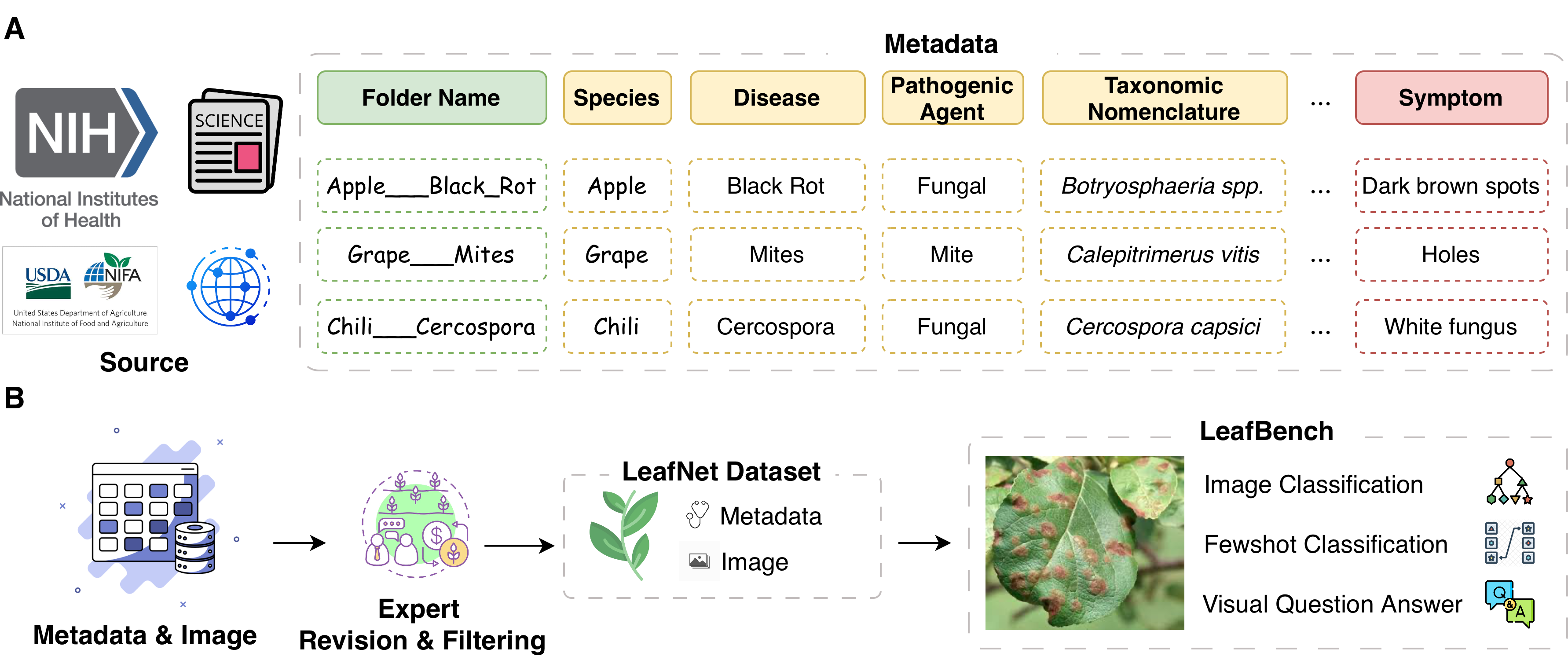}
    \caption{
    The LeafNet Curation and Benchmarking Pipeline.
    (\textbf{A}), Curation process overview. Metadata is synthesized from authoritative sources (NIH, NIFA) to map raw images to biological taxonomies, including species, disease, pathogenic agent, and symptom descriptions.
    (\textbf{B}), Expert verification and benchmarking. All metadata-image pairs undergo agricultural expert review to filter noisy samples. From this verified data, we construct LeafBench, a curated subset designed to benchmark \gls{LVLMs} on image classification, few-shot learning, and \gls{VQA}.}
\label{fig:LeafNet_full_pipeline}
\end{figure}
\end{graphicalabstract}

\begin{highlights}
    \item Introduce LeafNet, a publicly available, large-scale, multimodal, and standardized plant disease dataset, comprising 186,000 digital leaf images of 22 common crop species and 62 diseases, accompanied by rich annotated metadata.
    \item Introduce LeafBench, a comprehensive benchmarking framework for multiple Visual Question Answering (\gls{VQA}) tasks in plant disease detection and characterization.
    \item Conduct a systematic experimental analysis of \gls{VLMs} performance for multiple tasks in plant pathology.
    \item Investigate the challenges of few-shot learning for specialized plant pathology tasks on the LeafNet dataset.
\end{highlights}

%% Keywords

\end{frontmatter}

% \linenumbers


\section{Introduction}
\label{sec:introduction}

According to the 2025 report of the United Nations Food and Agriculture Organization (FAO), food insecurity and poverty remain critical, affecting more than 638 million people in 2024 due to growing demand and insufficient supply \cite{cd6008en}. According to FAO, up to 40\% of global crop yields are lost annually to pests and pathogens, resulting in economic losses exceeding USD 200 billion. This underscores the urgent need for more advanced data-driven methods to support informed decision-making in disease management and crop protection \cite{cc2447en}. Digital agriculture plays a pivotal role in monitoring and managing agricultural pests and diseases through advanced data collection and analysis, thereby mitigating yield losses and enhancing global food security \cite{00539-x}. Recent studies have demonstrated the effectiveness of integrating digital technologies, including Internet of Things with \gls{DL} \cite{106034}, autonomous vehicle deployment driven by machine learning and computer vision \cite{100483}, and the application of computer vision–based \gls{DL} approaches \cite{11100-x} for precision disease management. Despite these advances, significant challenges remain due to environmental variability and the overlapping visual symptoms of different diseases, which hinder accurate identification. To overcome these limitations, our study introduces a large-scale digital image dataset and a language benchmark designed to support the development of robust \gls{VLMs}, enabling not only image-based disease classification but also multimodal linguistic integration for zero-shot learning and \gls{VQA}.

To date, most studies have focused on distinguishing diseased plants from healthy plants using digital leaf images by employing \gls{DL} models and accuracy enhancement techniques. These works often employ baseline models such as Xception, ResNet50, MobileNets, and DenseNets, but are generally confined to specific datasets limited to a single crop, a single disease, or a particular location. The lack of cross-crop and cross-disease evaluation \cite{V7269,7447-2_47,0131011} restricts their transferability and hinders their applicability at larger scales. To address this limitation, several works have curated and tested the models on larger and more diverse datasets, such as PlantVillage, PlantDoc, and PlantSeg, achieving high accuracy of up to 98–99\% \cite{01405108,3.ch013,57613}. However, their performance remains limited in terms of generalization when applied to new datasets collected from a new environment. To address this, various strategies have been used, including data augmentation \cite{12746}, explainable \gls{AI} (XAI) \cite{5040804}, and advanced \gls{DL} models \cite{3240100,110166}. However, despite promising results, these approaches are largely confined to image-only models, which limits their ability to accurately distinguish between diseases that exhibit similar visual symptoms on leaves. For example, diseases such as brown spot and rice blast are difficult to distinguish on leaf images in early stages because they both produce similar necrotic, brown-black lesions on rice leaves. This highlights a key limitation of current image-based classification studies—narrow scope and restricted ability to generalize across diseases and conditions.

Foundation models and multimodal data have been widely discussed in modern \gls{AI}, particularly in benchmarking and in developing domain-specific systems, such as those for agriculture. A large-scale \gls{AI} model trained on massive datasets that serves as the basis for various downstream applications. These models leverage the full potential of unlimited fine-tuning when resources such as images, text, and audio are available. However, these models also present significant challenges regarding evaluation, comparison, contextual understanding, bias, and interpretability \cite{yuan23b, Awais2025}. Meanwhile, multimodal datasets that integrate information from different modalities, such as texts, images, video, audio, or other sensor signals, can provide multidimensional perspectives, which enables diverse applications. Although multimodal datasets strongly support the development of foundation models, they present challenges in data collection and synchronized annotation, particularly in plant disease detection and classification tasks \cite{3649447, 1408843, 103352}.
%Moreover, existing benchmarks primarily focus on supervised image classification, without fully leveraging the potential of few-shot, zero-shot, and \gls{VQA} in agriculture. %For instance, Weyler et al. introduced a large UAV-captured dataset with pixel-wise annotations and established standards for crop–weed recognition; however, it remains limited in scope (crop/weed), faces difficulties in leaf segmentation at complex growth stages, requires labour-intensive annotation, and lacks geographic diversity \cite{3419548}. Similarly, Liu et al. addressed plant disease diagnosis using diverse symptom datasets by assigning weights to each loss for label–region pairs during weak supervision, but its evaluation standards remain insufficiently comprehensive \cite{3049334}. Li et al. provided an extensive overview of vision technologies for weeding robots, adding valuable experimental benchmarks; nevertheless, the dataset was small, accuracy remained low, and real-world deployment was challenging for tall crops and complex environments \cite{106880}. Lei et al. presented a systematic and comprehensive benchmark for agricultural segmentation, aggregating public datasets; yet, it was constrained to standard benchmarks and in-depth quantitative analyses \cite{10775-6}. While these studies represent important advances in agricultural benchmarks, none have fully exploited the strengths of the approaches proposed in this work.

Furthermore, few-shot and zero-shot learning enable the model to perform efficiently with limited sample sizes. The method is beneficial for classifying rare diseases from only a few training examples. For example, few-shot learning can learn well from 5 images per disease class through pre-training, meta-training, and fine-tuning; while zero-shot learning allows classification of classes absent from training by transferring knowledge from the source domain to the target domain. The techniques can also work for novel diseases not present in the training data, based on the semantics of the text descriptions. Plant disease identification and characterization using deep learning requires large amounts of data, which are costly and time-consuming to collect, particularly for novel or rare diseases. Thus, few-shot learning addresses these limitations by demonstrating high effectiveness while reducing reliance on large-scale datasets. The method has been evaluated on several datasets, including PlantVillage in 38-way 16-shot \cite{Jiang2025}, PlantDoc with only five images per class \cite{Rezaei2024}, and PlantSeg with only 1-5 examples \cite{Iqbal2025}, and achieved an overall classification accuracy of approximately 94\%, 90\%, and 96\%, respectively. Meanwhile, zero-shot learning enables model generalization without requiring target-specific training data, leveraging shared semantic and visual features across domains to support inference without retraining. For example, model performance on the PlantVillage dataset surpasses that of traditional methods, achieving over 75\% accuracy \cite{SatyaRajendraSingh2023}, while results on the PlantSeg dataset (IoU = 0.593, Dice = 0.712, and other metrics exceeding previous approaches \cite{Bao2025}) demonstrate its effectiveness in addressing data scarcity. Given these advantages, this study employs few-shot and zero-shot learning as potential approaches to evaluate assumptions related to missing data in plant disease classification.

Developing robust, scalable deep learning models for disease detection and characterization requires a new, large-scale, multi-class dataset collected from diverse sources, encompassing a wide variety of crops and diseases. Here, we introduce LeafNet, a novel large-scale dataset comprising both RGB digital images and associated text annotations across multiple aspects (large-scale, multimodal, and standardized) to establish a foundation for the development of \gls{VLMs} in digital plant disease research and application. This facilitates diverse research directions, including few-shot and zero-shot learning, cross-modal retrieval, \gls{VQA}, and future advancements in domain adaptation and smart disease management. Alongside LeafNet, we developed the LeafBench dataset—a comprehensive evaluation resource designed to assess the progress of \gls{VLMs} in understanding plant pathology and to identify areas where \gls{VLMs} is limited in characterizing plant-pathogen interactions. The dataset and benchmark are expected to significantly accelerate the advancements of \gls{VLMs} in plant disease understanding, as illustrated in Figure \ref{fig:vlm_all_avg}. Overall, the main contributions of this study are as follows.
\begin{itemize}
    \item Introduce LeafNet, a publicly available, large-scale, multimodal, and standardized plant disease dataset, comprising 186,000 digital leaf images of 22 common crop species and 62 diseases, accompanied by rich annotated metadata.
    \item Introduce LeafBench, a comprehensive benchmarking framework for multiple \gls{VQA} tasks in plant disease detection and characterization.
    \item Conduct a systematic experimental analysis of \gls{VLMs} performance for multiple tasks in plant pathology.
    \item Investigate the challenges of few-shot learning for specialized plant pathology tasks on the LeafNet dataset.
\end{itemize}

% This study is organized into six sections. Section \ref{relatedwork} reviews related studies on multimodal learning and \gls{VLMs} and highlights the challenges of constructing benchmarks and datasets for plant disease research and application. Section \ref{methodology} describes the methodology used to build the LeafNet and LeafBench datasets for multiple vision-language tasks, with a particular focus on \gls{VLMs}. Section \ref{Experiments} outlines the experimental setup, implementation of various \gls{VLMs} and vision models, and results from the classification, few-shot, and zero-shot Visual Question Answering tasks. Finally, Section \ref{Conclusion} summarizes the key findings from the evaluation of the proposed datasets, Section \ref{Limitations} discusses current limitations, and Section \ref{future} outlines potential directions for future research.

This paper is organized into six sections. Section \ref{relatedwork} reviews related work on multimodal learning and \gls{VLMs}, highlighting challenges in constructing benchmarks and datasets for plant disease applications. Section \ref{methodology} describes the methodology for building LeafNet and LeafBench, focusing on dataset curation, task formulation for \gls{VLMs}, and experimental setup. Section \ref{Results} presents comprehensive evaluation results and in-depth analysis across zero-shot, few-shot, and fine-tuned settings, including comparative analysis between vision-only models and \gls{VLMs}. Section \ref{Conclusion} summarizes key findings from the benchmark evaluation and discusses current limitations. Finally, Section \ref{future} outlines directions for future research.

\section{Related Work}\label{relatedwork}

\subsection{Image-Based Machine Learning for Disease Characterization}

Machine learning-based plant disease classification is a crucial task in smart agriculture, where various machine learning methods are applied with varying degrees of success. First, image classification is applied in many disease identification tasks and has achieved promising results, such as studies on rice disease classification \cite{0131011}, grape leaf disease \cite{Vo2024}, tomato leaf disease \cite{ThaiNghe2023}, rice pests and diseases \cite{3355}, or studies that used large datasets such as PlantVillage \cite{10119138}. However, these models are limited by input image size, typically $224 \times 224$, $227 \times 227$, $256 \times 256$. This limitation can cause small, disease-specific features—such as early-stage rice blast on rice, black rot on grapes, or leaf blight on tomatoes—to be overlooked. Moreover, existing benchmarks mainly focus on supervised image classification, without fully leveraging the potential of few-shot, zero-shot, and \gls{VQA} in characterizing diseases. For example, Liu et al. addressed plant disease diagnosis using diverse symptom datasets by assigning weights to each loss for label–region pairs during weak supervision, but its evaluation standards remain inevitably comprehensive \cite{3049334}. Lei et al. presented a systematic and comprehensive benchmark for agricultural segmentation, aggregating public datasets; yet, it was constrained to standard benchmarks and in-depth quantitative analyses \cite{10775-6}. In general, machine learning methods or benchmark approaches have overcome the shortcomings of existing machine learning models. However, there are limitations in the number and criteria for building large and rich datasets to serve as standards for classifying leaf diseases.

\subsection{Multimodal and Vision-Language Models}

Over the past decade, foundation models have evolved to integrate multiple data for different modalities, particularly images and texts. Models such as GPT-4o \cite{gpt4o}, Gemini 2.5 Pro \cite{gemini}, and InternVL3.5 \cite{internvl3} have demonstrated the power of omni-modal and general-purpose assistants in processing diverse modalities (i.e., texts, images, audios, videos), moving toward \gls{AI} assistants with strong reasoning capabilities. Furthermore, models like CLIP \cite{clip}, ALIGN \cite{align}, SigLIP \cite{siglip}, and SigLIP2 \cite{siglip2} further demonstrate the effectiveness of contrastive learning, where image–text pairs are mapped into a shared representation space to enable retrieval and zero-shot classification. This learning technique establishes the foundation for various downstream tasks. More recently, large-scale models such as LLaVA 1.5 \cite{llava15}, LLaVA-NeXT \cite{llavanext}, Qwen2.5-VL \cite{qwen25}, and BLIP-2 \cite{blip-2} have propelled the development of instruction-tuned multimodal \gls{LLMs} by integrating visual encoders with \gls{LLMs} to enhance instruction-following \gls{VQA}, and image captioning capabilities. Domain-specific multimodal models like SCOLD \cite{scold} and BioCLIP \cite{bioclip} have demonstrated the potential of fine-tuning or customizing models to specific domains such as agriculture and biomedicine. Despite these achievements, most of these models are still largely trained on web-scale, general-purpose data, limiting their effectiveness in highly specialized contexts like biology or agriculture.

\subsection{Plant Disease Image Datasets and Benchmarks}

Over the past years, numerous plant disease image datasets have been developed to support the development of computer vision models for precision disease management. PlantVillage, for example, is one of the most popular and classical datasets for leaf disease classification, often used as a starting benchmark. However, it lacks diversity in field conditions and does not reflect the complexity of varying environments \cite{mohanty2016plantvillage}. In contrast, PlantDoc, PlantCLEF, and CropDeep address these limitations by acquiring data in real-world environments, offering more diverse information for detection and classification tasks; however, they often suffer from high noise due to the use of open-source data collection \cite{PlantDoc, PlantCLEF, CropDeep}. Other datasets, such as PhenoBench \cite{3419548}, FruitNet \cite{FruitNet}, and PlantSeg \cite{PlantSeg}, focus on specific crops, imaging conditions, or are designed primarily for morphological analysis (e.g., UAV imagery, fruit images, phenotyping). While these datasets have been valuable for image classification studies, they often lack diverse annotations and comprehensive metadata, which limits their applicability to more complex tasks, such as zero-shot learning or cross-modal reasoning.

Beyond individual datasets, several recent efforts have introduced benchmark frameworks to standardize model evaluation in agriculture. Agri-LLaVA proposed a benchmarking framework with fine-grained analysis to assess diversity and challenges in agricultural pests and pathogens, aiming to develop multimodal assistants for informed decision making \cite{agrillava}. AgroGPT built a customized vision–language model with multi-turn conversational capabilities, fine-tuning an expert \gls{LLM} on a dataset of 70,000 multimodal dialogues \cite{agrogpt}. \gls{CDDM} built a multimodal dataset for plant disease diagnosis to overcome poor generalization in agriculture. The dataset comprises 137,000 crop images, spanning 16 crop types and 60 disease classes, and about one million question–answer pairs designed for disease prevention and treatment strategies \cite{cddm}. AgriBench introduced the first agriculture-specific benchmark to evaluate the understanding and reasoning abilities of multimodal LLMs using a hierarchical benchmark approach, covering tasks from basic crop identification to agricultural strategy recommendation \cite{agrobench}. Similarly, Agri-342K developed AgriGPT, an ecosystem of agriculture-specific \gls{LLM}s, with a fine-tuning pipeline comprising pretraining → domain adaptation → instruction tuning → evaluation. This pipeline aimed at supporting farmers and experts in policy-making through \gls{AI} systems with domain-specific knowledge. However, this work focuses primarily on textual data and still requires further empirical validation \cite{agrigpt}. Despite these significant advances, most existing studies are still limited in terms of dataset size and task diversity. Thus, a standardized and comprehensive benchmark for the agricultural research community, especially for plant disease studies, has yet to be established.

To address the aforementioned limitations, here we introduces LeafNet and LeafBench, a large-scale multimodal dataset designed to accelerate the development of reliable vision-language models for plant disease identification, characterizing, reasoning, and disease management. The dataset and benchmark comprise more than 186,000 images covering 22 common crop species and 62 diseases, accompanied by rich textual annotations, including species, disease, pathogenic agent, symptom description, leaf status, and imaging conditions. Details of the dataset, along with comparisons to existing datasets are shown in Table \ref{tab:dataset_comparison_complete}. Building upon this resource, we developed a comprehensive benchmarking framework to evaluate the disease characterization performance of \gls{VLMs} across multiple settings: supervised classification, few-shot learning, zero-shot learning, and \gls{VQA}.

\begin{table*}[!htbp]
\centering
\footnotesize
\caption{Comparison of leaf disease datasets for vision-language model development.}
\label{tab:dataset_comparison_complete}
\renewcommand{\arraystretch}{1.3}
\begin{tabular}{|l|p{2cm}|c|p{2.5cm}|c|c|}
\hline
\textbf{Dataset} & \textbf{Annotation} & \textbf{Images} & \textbf{Image-Caption Pairs} & \textbf{QA Pairs} & \textbf{Main Purpose} \\
\hline
Agri-LLaVA \cite{agrillava} & GPT-4o \cite{gpt4o} & 391k & - & 68K (Synthetic) & Training \\ \hline
AgroInstruct \cite{agrogpt} & MedLLaVA \cite{llavamed} + Mistral \cite{mistral}  & 10k & 10k (Synthetic) & 70k (Synthetic) & Training \\  \hline
\gls{CDDM} \cite{cddm} & GPT-4 \cite{gpt4} & 137k & - & 1M (Synthetic) & Training \\ \hline
Agri-342K \cite{agrigpt} & Multi-agents & 342K & -- & 342K (Synthetic) & Training \\ \hline
AgriBench \cite{AgriBench}& Multi-agents & Unknown & -- & 13K (Synthetic) & Evaluation \\ \hline
AgroBench \cite{agrobench}& Expert & 3,745 & -- & 4,342 (Expert) & Evaluation \\ \hline
LeafNet (Ours) & Expert & 186K & 186K & -- & Training\\  \hline
LeafBench (Ours) & Expert & 2,910 & - & 13,950 (Expert) & Evaluation\\ 
\hline
\end{tabular}
\end{table*}

\section{Methodology} \label{methodology}

\subsection{LeafNet Overview}
The urgency of foundation model development in agriculture stems from their ability to address diverse tasks and domains across large-scale datasets \cite{espejo2025foundation}. To train such models effectively, the underlying data must be both sufficiently large and representative to establish a robust, domain-specific foundation. Conventionally, the PlantVillage \cite{mohanty2016plantvillage} dataset has served as the standard resource for leaf disease identification. Comprising over 54,000 images across 38 classes, the dataset has facilitated nearly 5,000 studies. However, its simplicity, marked by uniform backgrounds and limited biological diversity, makes it insufficiently challenging for modern \gls{DL} models \cite{richter2025benchmark,li2021plant,sarkar2023leaf}. Furthermore, the narrow scope of the dataset (14 crops, 26 diseases) leads to overfitting, hindering effective generalization to real-world agricultural scenarios \cite{PlantDoc}.

To bridge this gap, we developed LeafNet, which comprises over 186,000 images covering 22 economically significant crops and 62 disease categories. Uniquely, LeafNet aggregates data from multiple sources to ensure high intra-class variation and environmental diversity. To advance beyond basic classification, we developed LeafBench, a task-specific benchmark designed to integrate detailed disease and symptom descriptions into the model’s learning process. Rather than serving as a purely evaluative tool, LeafBench is central to our framework’s ability to improve identification accuracy and diagnostic reliability. Detailed construction and specifications are provided in Section \ref{sec:leaf_bench}.

\subsubsection{Curation Pipeline and Metadata}
To ensure that our dataset reflects both real-world precision and descriptive depth, we established a rigorous curation pipeline (Figure \ref{fig:LeafNet_full_pipeline}a). We prioritized biological accuracy by verifying taxonomic nomenclature and pathogen classifications against authoritative sources, such as the NIFA and NIH. To provide the semantic richness necessary for vision-language alignment, we extended standard labels with structured metadata (Table \ref{tab:metadata_descriptions}), incorporating detailed symptom descriptors like lesion morphology and chlorosis patterns.

\begin{figure}[H]
 \centering
 \includegraphics[width=1\textwidth]{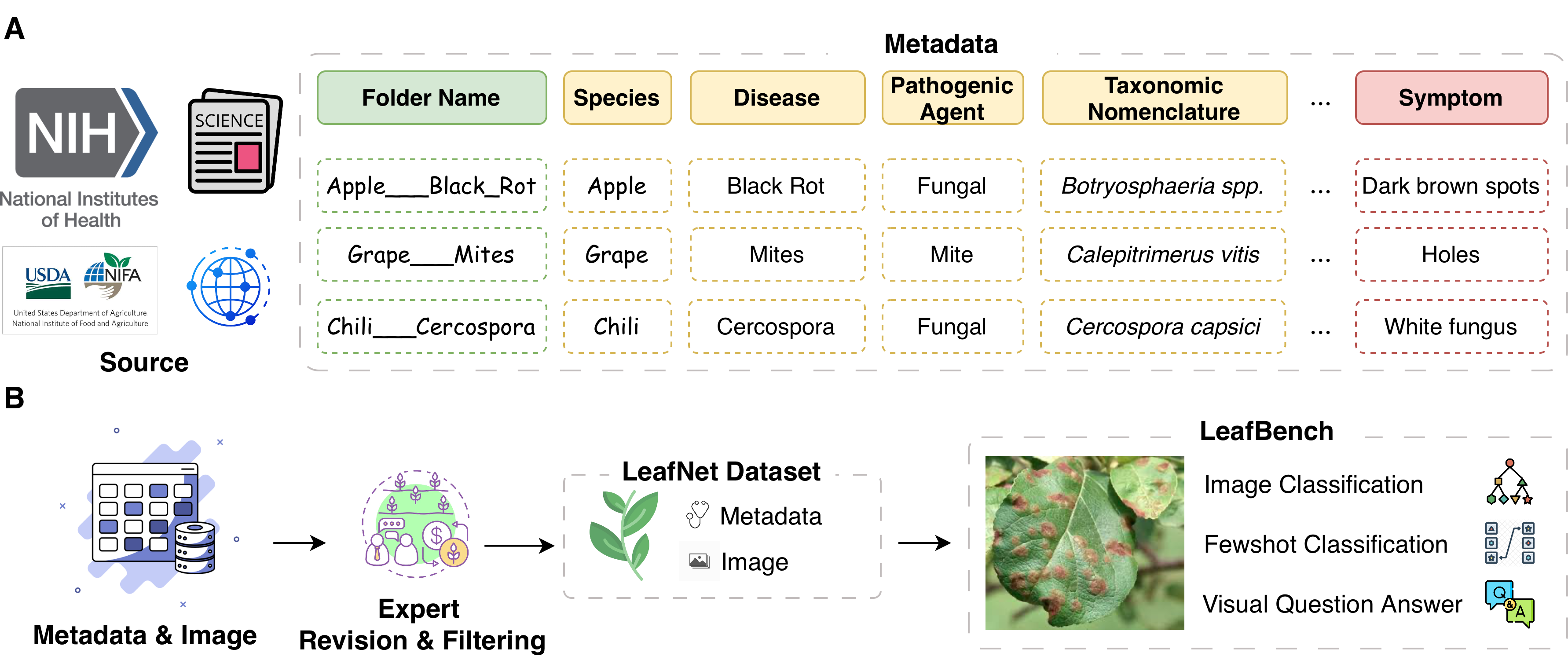}
    \caption{
    The LeafNet curation and benchmarking pipeline.
    (\textbf{A}), Curation process overview. Metadata is synthesized from authoritative sources (NIH, NIFA) to map raw images to biological taxonomies, including Species, Disease, Pathogenic Agent, and Symptom descriptions.
    (\textbf{B}), Expert verification and benchmarking. All metadata-image pairs undergo review by agricultural experts to filter out noisy samples. From this verified data, we construct \textbf{LeafBench}, a curated subset designed to benchmark \gls{LVLMs} on image classification, few-shot learning, and \gls{VQA}.}
\label{fig:LeafNet_full_pipeline}
\end{figure}

This metadata generation was followed by a human-in-the-loop validation process in which agricultural experts reviewed samples to filter mislabeled data and ensure that described symptoms were visually verifiable. This results in rich multimodal annotations that enable the training of domain-specific foundation models for accurate diagnosis and decision support.

\begin{table}[H]
\centering
\footnotesize
\caption{LeafNet dataset metadata descriptions and schema.}
\label{tab:metadata_descriptions}
\renewcommand{\arraystretch}{1.4} 
\begin{tabular}{|p{120pt}|p{320pt}|}
\hline
\textbf{Field} & \textbf{Description} \\
\hline
Folder Name & The standardized directory identifier serving as the ground truth label (e.g., Apple\_\_\_Black\_Rot). \\
\hline
Species & The biological classification of the host plant (e.g., \textit{Malus domestica} [Apple]). \\
\hline
Disease & The common agricultural name of the pathology (e.g., Black Rot, Rust). \\
\hline
Pathogenic Agent & The high-level category of the causal organism: Fungi, Bacterial, Viral, Oomycetes, or Mite. \\
\hline
Taxonomic Nomenclature & The formal scientific binomial of the specific pathogen (e.g., \textit{Botryosphaeria spp.}), complying with international taxonomic codes. \\
\hline
Sample Size & The quantitative count of valid image instances present within the specific class. \\
\hline
Symptom Description & A qualitative analysis of visual indicators, including lesion morphology, necrosis distribution, and colorimetric changes. \\
\hline
\multicolumn{2}{|l|}{\textbf{Technical Specifications}} \\
\hline
Image Format & Standard RGB color space (JPG encoding). \\
\hline
Resolution & Spatial dimensions (Height $\times$ Width) indicating feature granularity. \\
\hline
Acquisition Environment & Distinction between controlled laboratory settings and in-the-wild field conditions. \\
\hline
\end{tabular}
\end{table}

\subsubsection{LeafNet Details}
\begin{figure}[H]
    \centering
    \includegraphics[width=0.75\linewidth]{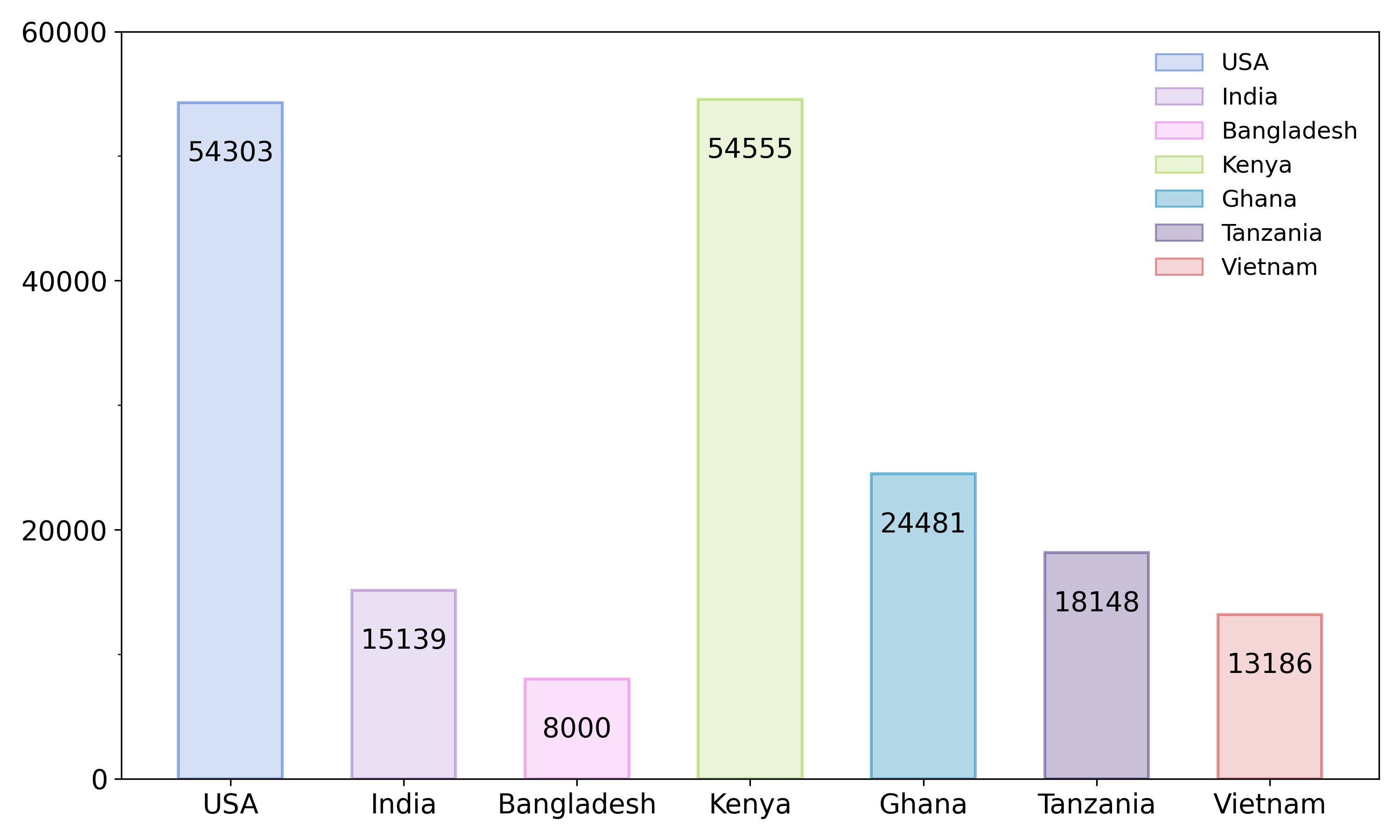}
    \caption{ The distribution of digital image acquisition by country.}
    \label{fig:country}
\end{figure}

Following the curation pipeline, we collected over 186,000 images with corresponding metadata from both public sources and the in-house collection. These data were collected from seven countries, with the sample distribution is illustrated in Figure \ref{fig:country}. 

The dataset features a broad geographic distribution, with major contributions from the USA and India, and covers a wide range of host plants, with coffee (\textit{Coffea}) and maize (\textit{Zea mays}) representing the most abundant classes (Table \ref{tab:crop_distribution}). 

\begin{table}[H]
\centering
\footnotesize
\caption{Detailed distribution of crop images in the LeafNet dataset.}
\label{tab:crop_distribution}
\renewcommand{\arraystretch}{1.3} 
\begin{tabular}{|l|c|l|c|l|c|}
\hline
\textbf{Crop} & \textbf{Count} & \textbf{Crop} & \textbf{Count} & \textbf{Crop} & \textbf{Count} \\
\hline
Coffee    & 49,375 & Cashew  & 5,777 & Potato     & 2,374 \\
\hline
Maize     & 27,744 & Orange  & 5,507 & Squash     & 1,964 \\
\hline
Tomato    & 25,838 & Soybean & 5,152 & Cherry     & 1,906 \\
\hline
Rice      & 13,107 & Mango   & 3,650 & Strawberry & 1,750 \\
\hline
Apple     & 12,458 & Peach   & 2,657 & Chili Pepper & 819 \\
\hline
Cassava   & 7,508  & Pepper  & 2,603 & Cucumber   & 800   \\
\hline
Sugarcane & 6,748  & Tea     & 2,520 & Raspberry  & 371   \\
\hline
Grape     & 5,867  & \multicolumn{2}{c}{} & \multicolumn{2}{c|}{} \\ 
\hline
\end{tabular}
\end{table}

Containing over 137,000 diseased leaves within a 186,000-image dataset, LeafNet offers significantly greater scale and diversity than existing datasets (Figure \ref{fig:dh_farm_lab}b).Previous studies relied predominantly on laboratory-acquired images characterized by uniform backgrounds and isolated leaf specimens. Although these models perform well on internal benchmarks, training on controlled-environment datasets often leads to significant performance drops on out-of-distribution samples. To mitigate this generalization gap, LeafNet emphasizes in-situ data acquisition, with images captured directly from cultivation sites (farms) comprising the majority of the dataset (Figure \ref{fig:dh_farm_lab}).

The word clouds in Figure \ref{fig:LeafNet_wordclouds} illustrate the diversity of symptoms and pathogenic agents represented in LeafNet. The distribution of host plants (Figure \ref{fig:LeafNet_wordclouds}a) indicates a balanced coverage across major agricultural crops, with 'Sugarcane', 'Tomato', and 'Rice' appearing prominently alongside 'Coffee'. From a pathological perspective, the dataset encompasses a broad biological range: as illustrated in Figure \ref{fig:LeafNet_wordclouds}c, 'Fungal' and 'Bacterial' pathogens dominate, while 'Viral' agents and 'Mite' vectors are also well represented. This coarse-grained categorization is further refined through detailed taxonomic classification (Figure \ref{fig:LeafNet_wordclouds}d), where scientific genera such as \textit{Xanthomonas} and \textit{Cercospora} occur frequently. Furthermore, the richness of the natural language descriptions is evident in the symptom analysis (Figure \ref{fig:LeafNet_wordclouds}e–f). These distributions show a heavy reliance on fine-grained visual descriptors—such as 'brown', 'yellow', 'spots', and 'lesions'. Collectively, these distributions underscore the dataset’s suitability for training vision–language models to align complex morphological patterns with textual descriptions.

\subsection{LeafBench Detail}
\label{sec:leaf_bench}
To facilitate a granular evaluation of diagnostic reliability, we curated LeafBench, a targeted benchmark derived from the LeafNet corpus. The full benchmark (\textit{All}) comprises 2,570 images and 13,950 samples, while a \textit{Tiny} subset (890 samples) enables cost-effective benchmarking for API-based foundation models and establishes a human expert baseline. 

Instead of relying on open-ended generation, which is prone to hallucination, LeafBench utilizes a {label-constrained prompting strategy. This approach requires \gls{VLMs} to perform fine-grained morphological analysis by selecting from a closed-world candidate set (e.g., Figure \ref{fig:qa_type}). For instance, in symptom identification task, the model must distinguish between visually similar features such as ``spots'' versus ``pustules'', aligning the model's visual reasoning with clinical diagnostic standards to improve overall classification accuracy.

This structured design evaluates model performance across six hierarchical diagnostic tasks:
\begin{itemize}
    \item \textbf{Disease \& pathogen identification:} Targets specific pathological conditions (DI) and high-level causal agents (PC) (e.g., Fungal vs. Viral).
    \item \textbf{Symptom identification (SI):} Focuses on fine-grained manifestations like lesion morphology and chlorosis to enhance diagnostic reliability.
    \item \textbf{Taxonomic \& species recognition:} Evaluates host plant species identification (CSI) across 22 species and binomial-level scientific naming (SNC).
    \item \textbf{Healthy-diseased classification (HDC):} Provides a fundamental binary assessment of the presence of pathology.
\end{itemize}

\begin{landscape}
\begin{figure*}[!ht]
 \centering
 \includegraphics[width=1.45\textwidth]{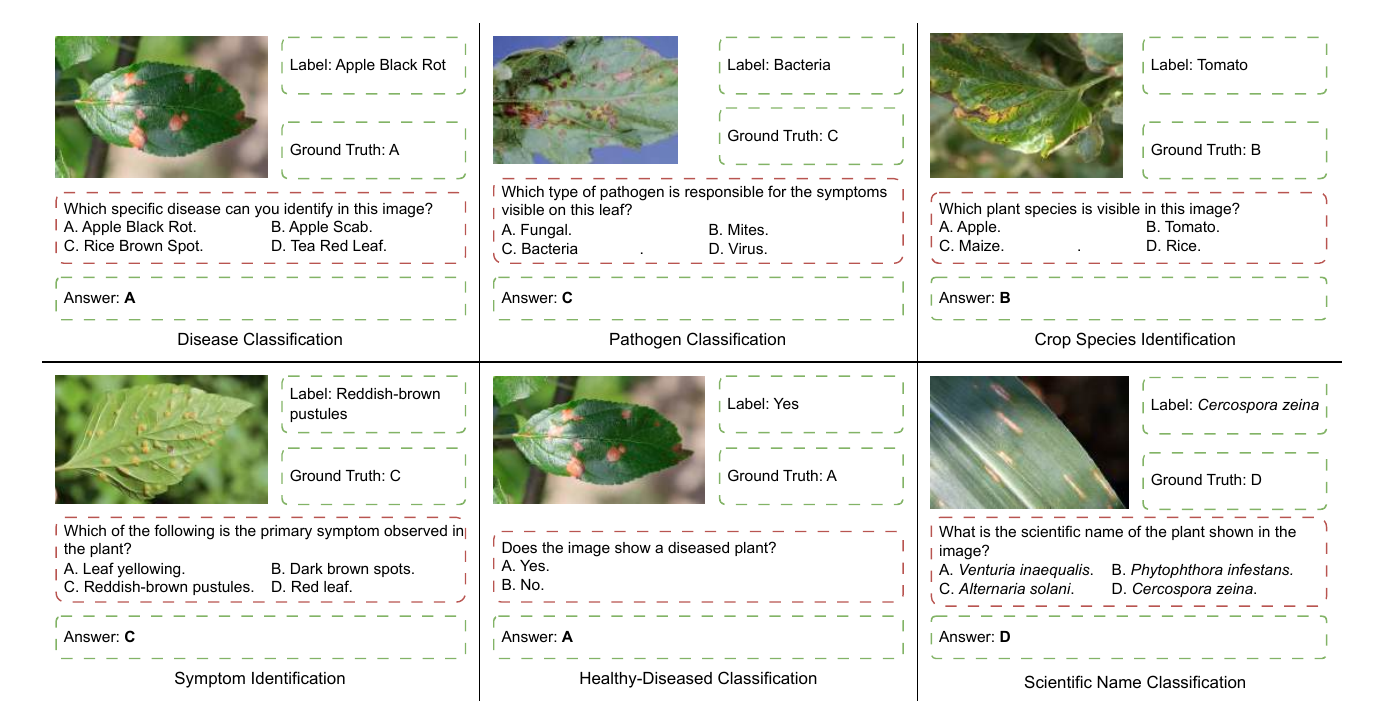}
    \caption{Illustration of the six Q\&A task types in LeafBench: Disease Identification (DI), Pathogen Classification (PC), Crop Species Identification (CSI), Symptom Identification (SI), Healthy–Diseased Classification (HDC), and Scientific Name Classification (SNC).}
\label{fig:qa_type}
\end{figure*}
\end{landscape}

By associating each task with a predefined set of candidate labels, the question prompts explicitly guide the model toward discriminative reasoning rather than unconstrained text generation. This design enables reliable output parsing, facilitates automated accuracy computation, and ensures consistent, context-bounded inference across LeafNet’s diverse biological hierarchy. Importantly, all questions are designed to be "visual-dependent", such that correct answers cannot be derived from the text prompt alone and instead require analysis of the accompanying image.

\begin{figure}[H]
    \centering
    \includegraphics[width=0.75\linewidth]{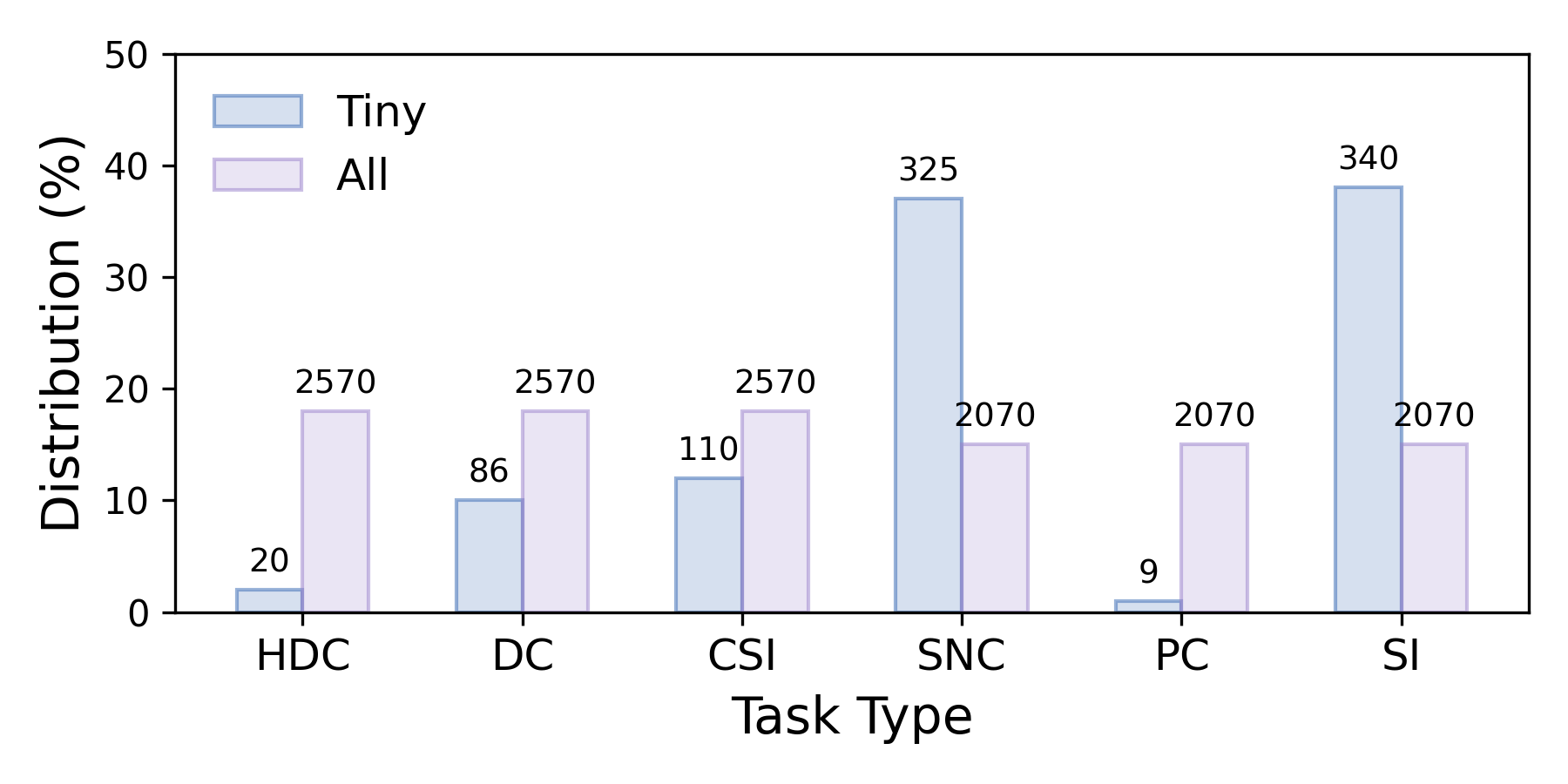}
    \caption{Distribution of question-answer pairs across diagnostic tasks in LeafBench. The bar chart shows the distribution for each of the six tasks: \textbf{Healthy-Diseased Classification (HDC)}, \textbf{Disease Classification (DC)}, \textbf{Crop Species Identification (CSI)}, \textbf{Scientific Name Classification (SNC)}, \textbf{Pathogen Classification (PC)}, and \textbf{Symptom Identification (SI)}. The numbers above the bars indicate the raw sample count for each category.}
    \label{fig:task_dist}
\end{figure}

The complexity of the prompting strategy is analyzed in Figure \ref{fig:len_dist}, which presents the distribution of token and character lengths across the six diagnostic tasks. Notably, the SI and SNC tasks exhibit the highest lexical density and variance (Figure \ref{fig:len_dist}b). This reflects the inherent difficulty of these tasks, which require the model to process detailed morphological descriptions and complex binomial nomenclature. In contrast, the binary HDC maintains a concise and uniform prompt structure, consistent with its high-level screening objective.

\begin{figure}[H]
    \centering
    \includegraphics[width=1\linewidth]{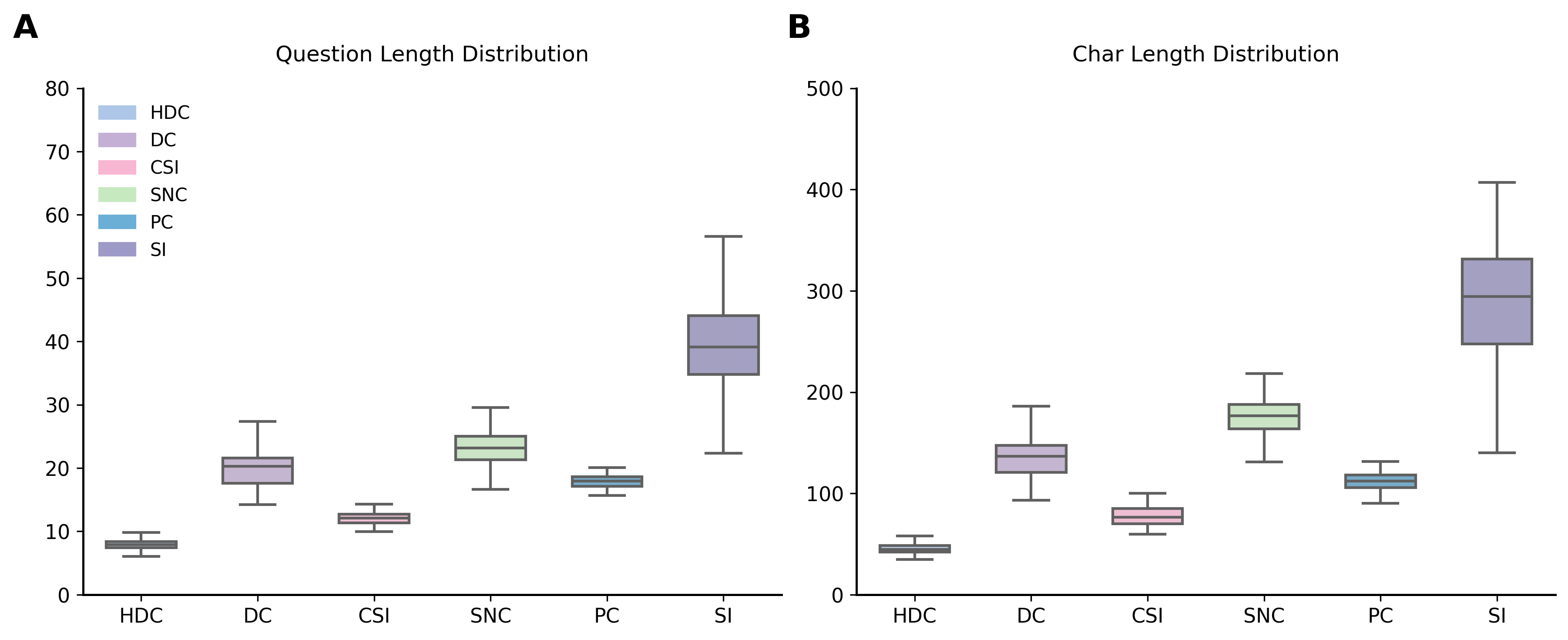}
    \caption{Quantitative analysis of prompt complexity in LeafBench. The distributions of (\textbf{A}) question lengths and (\textbf{B}) total character lengths are visualized across the six diagnostic tasks. The box plots reveal that fine-grained tasks, particularly Symptom Identification (SI) and Scientific Name Classification (SNC), require significantly richer textual context and exhibit higher variance compared to the concise prompts used for binary Healthy-Diseased Classification (HDC).}
\label{fig:len_dist}
\end{figure}

Complementing this structural analysis, Figure \ref{fig:answer_dist} visualizes the semantic diversity of the candidate answer space. The word clouds illustrate a comprehensive coverage of agricultural concepts, ranging from major crop commodities (Figure \ref{fig:answer_dist}a; e.g., `Apple', `Pepper') to high-level pathogenic categories (Figure \ref{fig:answer_dist}c; e.g., `Fungal', `Bacterial'). Crucially, the distribution of symptom descriptions (Figure \ref{fig:answer_dist}d) highlights a rich vocabulary of visual descriptors—such as `brown', `yellow', `spot', and `lesions'—ensuring that LeafBench rigorously evaluates the model's ability to ground fine-grained visual features to precise textual attributes.

\subsection{Experimental Protocols and Evaluation Methodology}

\subsubsection{Evaluation Strategy}
To assess the performance of vision and vision-language models on our LeafNet and LeafBench datasets, we established three rigorous evaluation protocols including 1) Visual Recognition and Data Efficiency, 2) Zero-Shot Semantic Alignment, and 3) Diagnostic Reasoning (Figure \ref{fig:LeafNet_tasks}). These protocols are structured according to dataset specificity and task complexity to evaluate representation quality, data efficiency, and advanced semantic reasoning:

\begin{itemize}
    \item \textbf{Visual recognition and data efficiency:} Conducted on the comprehensive \textbf{LeafNet} dataset, this protocol employs both fully supervised and few-shot classification paradigms. It is designed to assess the robustness of the visual backbone in extracting discriminative features from large dataset, as well as its adaptability when labeled samples are scarce.
    \item \textbf{Zero-shot semantic alignment:} Evaluated specifically on the LeafBench for CLIP-based models. This protocol assesses the intrinsic alignment between visual features and clinical text embeddings. By utilizing the semantic descriptions of diseases, it measures the model's ability to generalize to agricultural concepts without explicit fine-tuning.
    \item \textbf{Diagnostic reasoning:} Using the structured instruction set of \textbf{LeafBench}, this \gls{VQA} protocol is designed to evaluate model's capacity for instruction-based models. It tests the ability to interpret complex clinical queries and generate reliable, label-constrained answers about symptoms, pathogens, and taxonomy.
\end{itemize}

\begin{figure*}[!ht]
 \centering
 \includegraphics[width=1.0\textwidth]{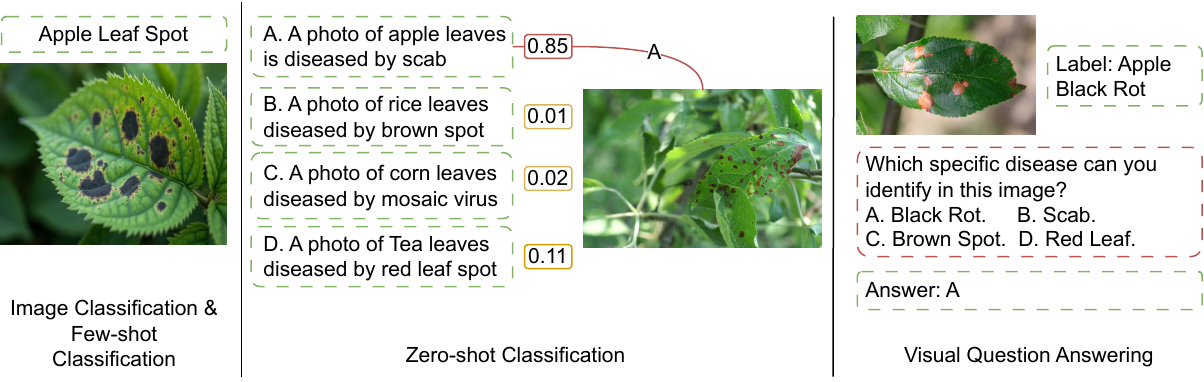}
    \caption{Overview of the three model evaluation protocols. (Left) Supervised and few-shot classification. (Center) zero-shot classification via image-text embedding alignment. (Right) \gls{VQA} tasks evaluating reasoning capabilities on the LeafBench subset.}
    \label{fig:LeafNet_tasks}
\end{figure*}

\subsubsection{Evaluation Tasks}

\textbf{Large-scale supervised classification—vision:} This task establishes the baseline diagnostic capability of vision models using the full LeafNet dataset. The objective is to classify leaf images into one of 97 fine-grained categories, encompassing 22 crop species, 62 specific diseases, and healthy control class. To decouple the quality of learned representations from the model's capacity to fit the training data, the model performance was evaluated under two distinct regimes:
\begin{itemize}
    \item \textbf{Linear probing:} The backbone features are frozen, and only the linear classification head is trained. This evaluates the robustness of the pre-trained features and their direct transferability to the plant disease domain without weight updates.
    \item \textbf{Full fine-tuning:} The entire model (backbone and classifier) is updated. This evaluates the maximum diagnostic potential of the architecture when fully adapted to the plant disease domain.
\end{itemize}

\textbf{Few-shot classification and transfer learning}: Agricultural environments frequently encounter "long-tail" distributions where data for emerging pathogens or rare physiological disorders are scarce \cite{110244}. To simulate these real-world constraints, we employ a few-shot classification protocol. This task assesses the model's ability to generalize to disease ontologies using limited supervision. We adopt an $N$-shot evaluation setting, where the model is trained on a randomly sampled support set containing $N$ examples per class, with $N \in \{16, 32, 64\}$. This logarithmic progression allows us to analyze the model's learning curve and determine the minimal data requirement for effective disease detection.

\textbf{Zero-shot visual question answering}:
Beyond simple classification, diagnostic support systems must possess reasoning capabilities. Using the LeafBench benchmark, this task evaluates the robustness of \gls{VLMs} in a zero-shot setting—meaning the models are not fine-tuned on the QA pairs. We assessed model performance on closed-form QA Multiple-choice questions requiring precise discrimination between confusingly similar symptoms (e.g., distinguishing \textit{Early Blight} from \textit{Late Blight} based on lesion morphology). This protocol tests the alignment between visual features and complex linguistic agricultural concepts, measuring the model's ability to perform expert-level reasoning without task-specific training.

\subsection{Experimental Settings}\label{Experiments}
\textbf{Baseline models:} For \gls{VLMs}, we use two close-source models: GPT-4o\cite{gpt4o} and Gemini 2.5 Pro\cite{gemini} as baselines for model training. For comparison, ten open-source models including LLaVA 1.5\cite{llava15}, Qwen 2.5 VL \cite{qwen25}, BLIP-2 \cite{blip-2}, CLIP \cite{clip}, BioCLIP \cite{bioclip}, SCOLD \cite{scold},SigLIP \cite{siglip}, SigLIP2 \cite{siglip2}, InternVL3.5 \cite{internvl3}, and ALIGN \cite{align} were evaluated. For vision models, we assessed seven models: VGG16 \cite{vgg}, Swin Transformer Tiny (SwinT) \cite{swin}, MobileNetV3 Small \cite{mobilenetv3}, Vision Transformer (ViT) \cite{vit}, EfficientNetB0 \cite{efficientnetb0}, EfficientNetV2 Small (EfficientNetV2S) \cite{efficientnetv2} and DenseNet121\cite{densenet}. Table~\ref{tab:models_summary} summarizes the parameter setup, pretrained domains, and architectures of all baseline models considered in our experiments.

\begin{landscape}
\begin{table*}[!ht]
\centering
\footnotesize
\caption{Vision and vision-language models used in this study.}
\label{tab:models_summary}
\begin{tabular}{|c|c|l|c|}
\toprule
\textbf{Model} & \textbf{Parameters} & \textbf{Pre-training Dataset} & \textbf{Domain} \\
\midrule
\multicolumn{4}{|c|}{\textit{General Large Multimodal Models}} \\
\midrule
GPT-4o \cite{gpt4o} & $\sim 200$B & Web crawl + LAION + Code + Math + Multimodal data & General \\  \hline
Gemini 2.5 Pro \cite{gemini} & $\sim 288$B & Web documents + Code + Images + Audio + Video & General \\ \hline
LLaVA 1.5 \cite{llava15} & $\sim 7$B & LAION-CC-SBU-558K + LLaVA-Instruct-665K & General \\ \hline
LLaVA-NeXT \cite{llavanext} & $\sim$7B & LAION-CC-SBU-558K + M4-Instruct-1177K & General \\ \hline
Qwen 2.5 VL \cite{qwen25} & $7$B & Web text + Images + Videos (Dynamic FPS) & General \\ \hline
BLIP-2 \cite{blip-2} & $\sim 2.7$B & LAION + CC3M + CC12M + SBU + VG & General \\ \hline
InternVL3.5 \cite{internvl3} & $\sim$8B  & Multimodal corpus (56M samples, 130B tokens) & General \\ \hline
SigLIP \cite{siglip} & $\sim 400$M & WebLI (English image-text pairs) & General \\ \hline
SigLIP2 \cite{siglip2} & $\sim 400$M & WebLI + Multilingual data + ACID curation & General \\ \hline
ALIGN \cite{align} & $\sim 1.8$B & Noisy alt-text pairs (1.8B image-text pairs) & General \\ \hline
CLIP \cite{clip} & $\sim 400$M & WebImageText-400M & General \\ 
\midrule
\multicolumn{4}{|c|}{\textit{Domain-specific Vision-Language Models}} \\ 
\midrule
BioCLIP \cite{bioclip} & $\sim 149$M & TreeOfLife-10M + iNat21 + BIOSCAN-1M + EOL & Biology \\
SCOLD \cite{scold} & $\sim 180$M & LeafNet-186K (Plant images + Symptom descriptions) & Plant Pathology \\
\midrule
\multicolumn{4}{|c|}{\textit{Vision Models}} \\
\midrule
VGG16 \cite{vgg} & $\sim 138$M & ImageNet-1K & General \\ \hline
SwinT \cite{swin} & $\sim 29$M & ImageNet-1K & General \\ \hline
MobileNetV3 Small \cite{mobilenetv3} & $\sim 2.54$M & ImageNet-1K & General \\ \hline
Vision Transformer & $\sim 86$M & ImageNet-21K $+$ ImageNet-1K & General \\ \hline
EfficientNetB0 \cite{efficientnetb0} & $\sim 5.3$M & ImageNet-1K & General \\ \hline
EfficientNetV2S \cite{efficientnetv2} & $\sim 21$M & ImageNet-21K & General \\ \hline
DenseNet121 \cite{densenet} & $\sim 8$M & ImageNet-1K & General \\
\bottomrule
\end{tabular}
\end{table*}
\end{landscape}

\textbf{Experiment setup for LeafNet}: LeafNet dataset is divided into training, validation, and test sets with a ratio of $6:2:2$. Vision models (which use only images as input) are trained under two regimes: fully fine-tuning all parameters and linear probing, in which only the classification layer is updated to assess the representational quality of pre-trained features. For the few-shot image classification, vision models are evaluated with $16$, $32$, and $64$ shots per class. Models are trained with both fully fine-tuning and linear probing to test their adaptability under data-scarce settings. For all experiments, we employ the AdamW optimizer \cite{Loshchilov} with a learning rate of $1\times10^{-3}$, weight decay of $0.05$, and a batch size of $32$. The hyperparameter configurations are summarized in Table~\ref{tab:hyperparameters}.

\textbf{Experiment setup for LeafBench}: For the zero-shot method, we evaluated CLIP-based models without task-specific fine-tuning. Models are required to align leaf images with each answer. Generative models were directly tested on six question-answer types about leaf images in LeafBench.

\textbf{Evaluation metrics}: For image classification and few-shot learning tasks on LeafNet, we employed a comprehensive set of evaluation metrics including: Accuracy (Acc), Precision, Recall, F1 Score, \gls{AUC}, and Cohen Kappa. For the \gls{VQA} tasks on LeafBench, we use Acc as the primary evaluation metric, measuring the percentage of correctly answered questions by comparing model predictions against ground-truth answers.

\section{Results and Discussion}\label{Results}

\subsection{Image Classification with Vision-Only Models}
To evaluate the utility of LeafNet as a foundation for agricultural representation learning, we benchmarked seven diverse vision architectures under both linear probing and full fine-tuning  protocols.

% \textbf{Dataset Quality and Learnability}: As detailed in Table \ref{tab:ft_performance}, the fine-tuning results serve as a primary validation of LeafNet's data quality. Across distinct architectural paradigms—ranging from CNN (e.g., DenseNet121, EfficientNet) to Vision Transformers (e.g., SwinT, ViT)—models consistently achieved high discriminatory performance, with \gls{AUC} scores exceeding 99\% and Cohen’s Kappa values surpassing 0.95. These metrics demonstrate that LeafNet contains rich, consistent, and learnable visual patterns, providing a sufficient signal-to-noise ratio to train high-precision diagnostic models. The leadership of DenseNet121 (94.27\% accuracy) further indicates that LeafNet benefits from feature reuse mechanisms inherent in dense connections. Furthermore, the strong performance and consistently high disease classification accuracy—approximately 90\% or higher—observed across all tested deep learning architectures demonstrate the robustness of our image dataset to a range of deep learning backbones.

\textbf{Dataset quality and learnability}: As detailed in Table \ref{tab:merged_performance}, the fine-tuning results serve as a primary validation of LeafNet's data quality. While all models achieved \gls{AUC} scores exceeding 99\%—indicating a high signal-to-noise ratio—we observe distinct performance stratifications driven by architectural inductive biases. Among the evaluated architectures, \textbf{DenseNet121} achieved the highest accuracy (94.27\%) and F1-score (94.12\%). This superior performance is likely attributed to its dense connectivity pattern, which promotes feature reuse and improves gradient flow. In the context of plant pathology, where distinguishing diseases often relies on fine-grained textures (e.g., fungal spots vs. bacterial lesions), DenseNet's ability to preserve low-level features throughout deep layers proves critical. Meanwhile, we observe a significant performance gap between SwinT (93.41\%) and ViT (89.72\%). While standard ViT's global processing struggles to resolve small necrotic lesions, SwinT's hierarchical shifted-window mechanism effectively captures these high-frequency details. By bridging CNN-like locality with long-range dependency modeling, SwinT achieves superior symptom delineation without the massive pre-training scale typically required by standard ViTs.

% \begin{table}[H]
%     \centering
%     \caption{Image classification performance on LeafNet test set under the fine-tuning setup. High metrics across diverse architectures confirm the dataset's consistency and learnability.}
%     \label{tab:ft_performance}
%     \begin{tabular}{|l|c|c|c|c|c|c|}
%         \toprule
%         \textbf{Model} & \textbf{Acc} & \textbf{Prec} & \textbf{Recall} & \textbf{F1} & \textbf{Kappa} & \textbf{\gls{AUC}} \\
%         \midrule
%         ViT & 89.72 & 90.07 & 89.72 & 89.52 & 95.08 & 99.86 \\ \hline
%         VGG16 & 90.99 & 91.63 & 90.99 & 91.09 & 95.32 & 99.85 \\ \hline
%         MobileNetV3Small & 89.44 & 89.74 & 89.44 & 89.21 & 96.15 & 99.88 \\ \hline
%         DenseNet121 & \textbf{94.27} & \textbf{94.46} & \textbf{94.27} & \textbf{94.12} & 98.51 & \textbf{99.96} \\ \hline
%         EfficientNetB0 & 93.31 & 93.53 & 93.31 & 93.25 & 97.09 & \underline{99.94} \\ \hline
%         EfficientNetV2S & \underline{93.47} & 93.76 & \underline{93.47} & \underline{93.41} & \textbf{98.03} & \underline{99.94} \\ \hline
%         SwinT & 93.41 & \underline{94.11} & 93.41 & 93.33 & \underline{98.01} & 99.95 \\ 
%         \bottomrule
%     \end{tabular}
% \end{table}

\textbf{Domain specificity and transfer challenges}: Conversely, the linear probing results (Table \ref{tab:merged_performance}) underscore the necessity of LeafNet as a specialized domain benchmark. We observe a significant ``generalization gap'' when relying solely on pre-trained ImageNet features, with accuracy drops ranging from 13.5\% to over 67\% compared to fine-tuning. This discrepancy highlights that visual symptoms and patterns in plant pathology are fundamentally distinct from general object recognition tasks, and that broadly pre-trained deep learning models are insufficient for this application. Notably, while older architectures like VGG16 maintain reasonable transferability (77.48\% accuracy), modern efficient architectures (e.g., EfficientNetV2S) performs poorly in the LP setting (25.79\%). This finding positions LeafNet as a critical resource for researching \emph{domain adaptation}, proving that generic visual features are insufficient for precision agriculture and highlighting the need for specialized, large-scale training data such as that provided by LeafNet.

% \begin{table}[H]
%     \centering
%     \caption{Image classification performance on LeafNet test set under the linear probing setup. The performance drop compared to FT highlights the unique, domain-specific distribution of the LeafNet dataset.}
%     \label{tab:lp_performance}
%     \begin{tabular}{|l|c|c|c|c|c|c|}
%         \toprule
%         \textbf{Model} & \textbf{Acc} & \textbf{Prec} & \textbf{Recall} & \textbf{F1} & \textbf{Kappa} & \textbf{\gls{AUC}} \\
%         \midrule
%         ViT & 45.01 & 38.84 & 45.01 & 38.72 & 48.79 & 93.76 \\ \hline
%         VGG16 & \textbf{77.48} & \textbf{78.36} & \textbf{77.48} & \textbf{76.72} & \textbf{87.34} & \textbf{99.44} \\ \hline
%         MobileNetV3Small & 44.37 & 37.98 & 44.37 & 38.96 & 49.32 & 92.97 \\ \hline
%         DenseNet121 & \underline{70.74} & \underline{70.13} & \underline{70.74} & \underline{67.99} & \underline{78.63} & \underline{98.54} \\ \hline
%         EfficientNetB0 & 27.25 & 20.43 & 27.25 & 18.86 & 24.29 & 82.35 \\ \hline
%         EfficientNetV2S & 25.79 & 19.05 & 25.79 & 15.06 & 10.92 & 81.39 \\ \hline
%         SwinT & 31.75 & 23.47 & 31.75 & 24.99 & 30.51 & 88.25 \\
%         \bottomrule
%     \end{tabular}
% \end{table}

\begin{landscape}
\small
\begin{table}[H]
    \centering
    \caption{Comparison of image classification performance on the LeafNet test set under fine-tuning and linear probing setups.}
    \label{tab:merged_performance}
    \resizebox{\linewidth}{!}{% Optional: Resizes table to fit width if necessary
    \begin{tabular}{l|cccccc|cccccc}
        \toprule
        \multirow{2}{*}{Model} & \multicolumn{6}{c|}{Fine-Tuning} & \multicolumn{6}{c}{Linear Probing } \\
        \cmidrule(lr){2-7} \cmidrule(lr){8-13}
         & Acc & Prec & Recall & F1 & Kappa & \gls{AUC} & Acc & Prec & Recall & F1 & Kappa & \gls{AUC} \\
        \midrule
        ViT \cite{vit} & 89.72 & 90.07 & 89.72 & 89.52 & 95.08 & 99.86 & 45.01 & 38.84 & 45.01 & 38.72 & 48.79 & 93.76 \\ \hline
        VGG16 \cite{vgg} & 90.99 & 91.63 & 90.99 & 91.09 & 95.32 & 99.85 & \textbf{77.48} & \textbf{78.36} & \textbf{77.48} & \textbf{76.72} & \textbf{87.34} & \textbf{99.44} \\ \hline
        MobileNetV3Small \cite{mobilenetv3} & 89.44 & 89.74 & 89.44 & 89.21 & 96.15 & 99.88 & 44.37 & 37.98 & 44.37 & 38.96 & 49.32 & 92.97 \\ \hline
        DenseNet121 \cite{densenet} & \textbf{94.27} & \textbf{94.46} & \textbf{94.27} & \textbf{94.12} & \textbf{98.51} & \textbf{99.96} & \underline{70.74} & \underline{70.13} & \underline{70.74} & \underline{67.99} & \underline{78.63} & \underline{98.54} \\ \hline
        EfficientNetB0 \cite{efficientnetb0} & 93.31 & 93.53 & 93.31 & 93.25 & 97.09 & \underline{99.94} & 27.25 & 20.43 & 27.25 & 18.86 & 24.29 & 82.35 \\ \hline
        EfficientNetV2S \cite{efficientnetv2} & \underline{93.47} & 93.76 & \underline{93.47} & \underline{93.41} & \underline{98.03} & \underline{99.94} & 25.79 & 19.05 & 25.79 & 15.06 & 10.92 & 81.39 \\ \hline
        SwinT \cite{swin} & 93.41 & \underline{94.11} & 93.41 & 93.33 & 98.01 & 99.95 & 31.75 & 23.47 & 31.75 & 24.99 & 30.51 & 88.25 \\ 
        \bottomrule
    \end{tabular}
    }
\end{table}
\end{landscape}

\subsection{Few-Shot Classification}

To evaluate data efficiency under label-scarce conditions, we benchmarked model performance across varying shot counts ($k \in \{16, 32, 64\}$) under two distinct regimes: full fine-tuning (Figure \ref{fig:fewshot_ft}) and linear probing (Figure \ref{fig:fewshot_lp}). Our results reveal critical trade-offs between architectural complexity and adaptability in the agricultural domain.

% \textbf{Fine-Tuning Efficiency}: Under the full fine-tuning protocol (Figure \ref{fig:fewshot_ft}), models exhibited rapid adaptation to the LeafNet domain. DenseNet121 consistently outperformed all other architectures, achieving an accuracy of $\approx$63\% at 16-shot and scaling robustly to over 82\% at 64-shot. This trajectory indicates that dense connectivity patterns facilitate effective feature reuse even with limited supervision. Similarly, the SwinT demonstrated strong scalability, reaching parity with DenseNet121 in the 64-shot regime ($\approx$79\% accuracy). In contrast, lighter architectures, such as MobileNetV3Small and EfficientNetB0, showed slower convergence, suggesting that their compact capacity may limit rapid adaptation when data are extremely scarce.

\textbf{Fine-tuning efficiency and architectural analysis}: Under the few-shot protocol (Figure \ref{fig:fewshot_ft}), we observe distinct performance stratifications driven by architectural inductive biases. DenseNet121 proved the most data-efficient ($\approx$63\% at 16-shot to $\approx$83\% at 64-shot), attributing its success to dense connectivity which maximizes feature reuse in scarce-data regimes. The SwinT showed comparable scalability ($\approx$79\% at 64-shot), leveraging hierarchical shifted windows to capture local disease morphology more effectively than standard global attention. Notably, ViT-B/16 outperformed legacy architectures like VGG16 by a wide margin ($\approx$20\% gap at 16-shot), indicating that modern pre-trained representations transfer better to plant pathology than deeper, non-residual CNNs which are prone to overfitting. Among lightweight models, EfficientNetV2S displayed robust scaling ($\approx$50\% to $\approx$75\%), significantly outpacing MobileNetV3Small, confirming that modern compound scaling preserves essential expressivity for rapid adaptation even with limited capacity.

\begin{figure*}[!ht]
 \centering
 \includegraphics[width=1.0\textwidth]{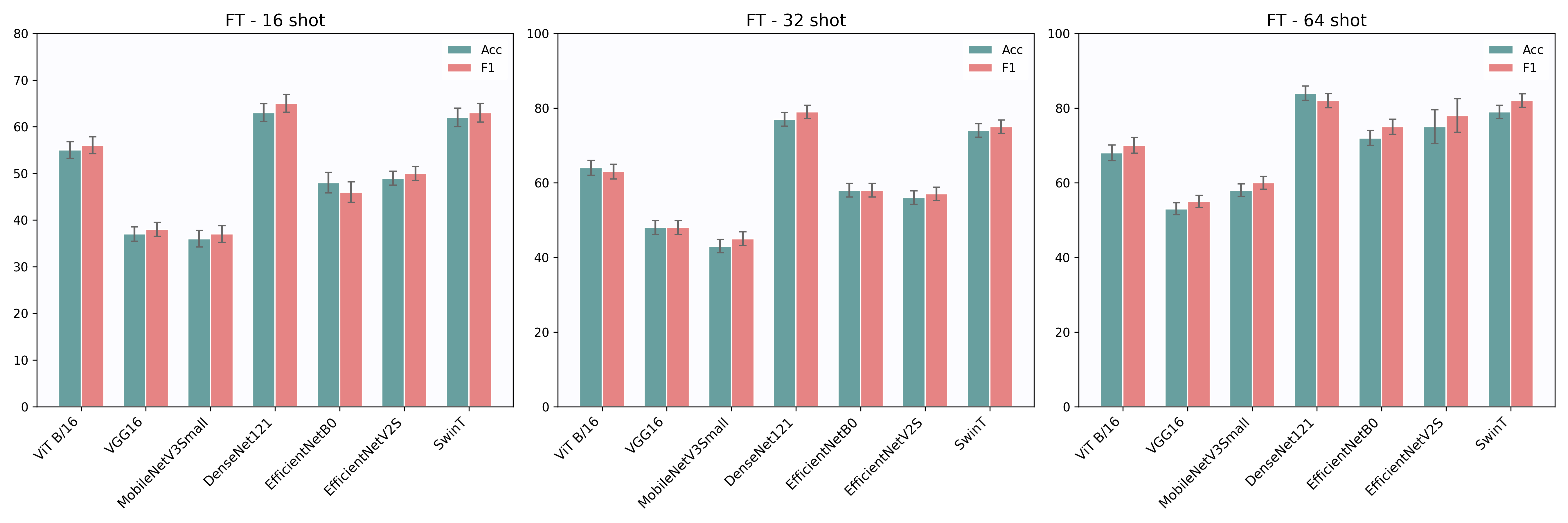}
    \caption{Few-shot classification performance on LeafNet across 16, 32, and 64-shot settings under fine-tuning setup. Error bars denote standard deviation across 10 random seeds.}
    \label{fig:fewshot_ft}
\end{figure*}

\textbf{Linear probing and feature robustness}: The linear probing results (Figure \ref{fig:fewshot_lp}) highlight a significant generalization gap. Without parameter updates to the backbone, performance drops precipitously across most modern architectures. For instance, EfficientNetV2S, which performs competitively in full fine-tuning, collapses to $<20\%$ accuracy in the LP setting, indicating that its ImageNet-pretrained features are highly specialized and misaligned with plant pathology features. Notably, however, VGG16 appears to be more robust in this fixed-feature regime, rivaling DenseNet121 with $\approx$65\% accuracy at 64-shot. This suggests that while older, over-parameterized CNNs may lack the peak efficiency of modern designs, their learned representations are surprisingly more transferable to out-of-distribution texture tasks like leaf disease identification.

\begin{figure*}[!ht]
 \centering
 \includegraphics[width=1.0\textwidth]{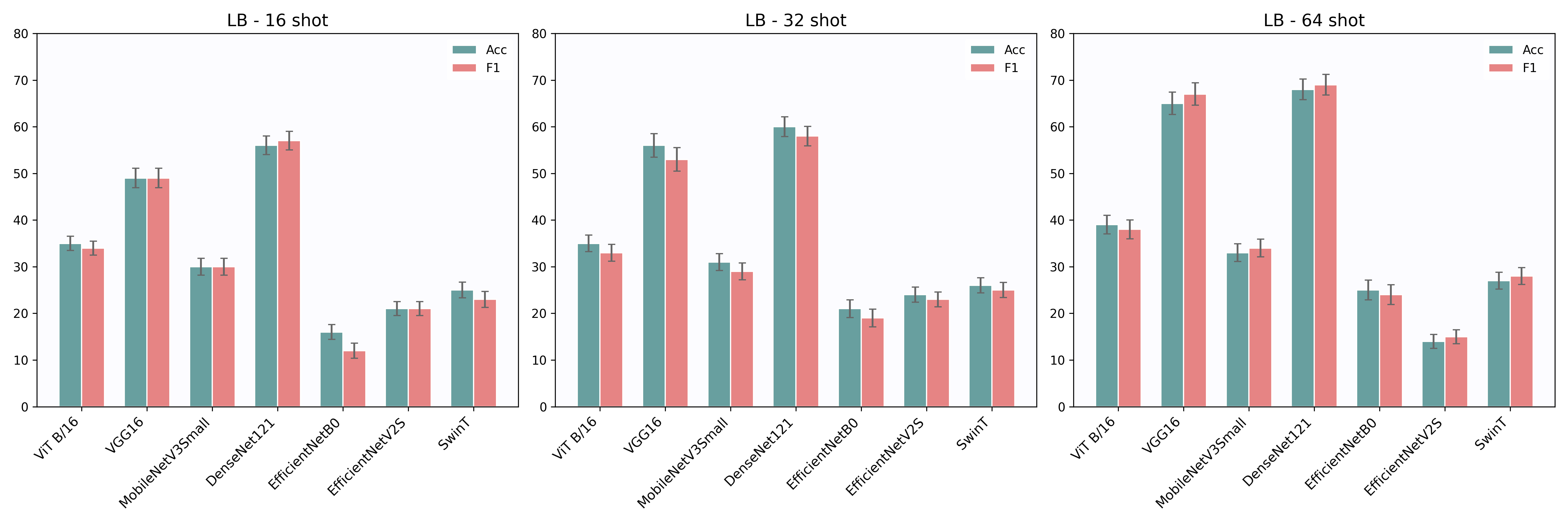}
    \caption{Few-shot classification performance on LeafNet across 16, 32, and 64-shot settings under linear pro. Error bars denote standard deviation across 10 random seeds.}
    \label{fig:fewshot_lp}
\end{figure*}

\subsection{Vision-Only vs Vision-Language Models}
To assess the potential of multi-modal architectures, we compared Vision-Only models against \gls{VLMs} using a $60:40$ split of the LeafBench dataset for fine-tuning and evaluation. As shown in Table \ref{tab:vision_vs_vlm_trained}, VLMs demonstrate a clear performance advantage.

While traditional vision models remain competitive in standard classification tasks such as HDC, achieving $94.27\%$ accuracy, they struggle to generalize to semantically demanding tasks like SNC and SI (e.g., $61.88\%$ and $59.22\%$ with DenseNet121 and SwinT, respectively). In contrast, fine-tuned VLMs demonstrate superior robustness across all categories. SCOLD achieves near-perfect disease identification ($99.15\%$) while maintaining substantial accuracy gains over the best vision models in semantic tasks ($+27.76\%$ and $+30.24\%$ for SNC; $+24.69\%$ and $+25.47\%$ for SI). These results confirm that integrating linguistic representations significantly enhances diagnostic precision compared to purely visual approaches.

\begin{table}[h]
    \small
    \centering
    \caption{Performance comparison between Vision-Only models and Vision-Language Models (VLMs) on the LeafBench benchmark.}
    \label{tab:vision_vs_vlm_trained}
    \resizebox{\linewidth}{!}{
    \begin{tabular}{l|l|cccccc}
        \toprule
        \textbf{Type} & \textbf{Model} & \textbf{HDC} & \textbf{DC} & \textbf{CSI} & \textbf{SNC} & \textbf{PC} & \textbf{SI} \\
        \midrule
        \multirow{4}{*}{Vision-Only} 
         & VGG16 \cite{vgg}  & 90.99 & 91.09 &
         85.12 & 54.30 & 88.75 & 66.41 \\
         & DenseNet121 \cite{densenet} & 94.27 & 94.12 & 92.45 & 61.88 & 93.50 & 70.23 \\
         & ViT \cite{vit} & 89.72 & 89.52 & 82.33 & 50.15 & 86.40 & 63.10 \\
         & SwinT \cite{swin} & 93.41 & 93.33 & 91.05 & 59.22 & 92.18 & 69.45 \\
        \midrule
        \multirow{2}{*}{VLM} 
         & CLIP \cite{clip} & \underline{97.50} & \underline{96.20} & \underline{94.10} & \underline{82.50} & \underline{95.80} & \underline{88.15} \\
         & SCOLD \cite{scold} & \textbf{99.15} & \textbf{98.85} & \textbf{96.73} & \textbf{89.64} & \textbf{97.83} & \textbf{94.92} \\
        \bottomrule
    \end{tabular}
    }
\end{table}

\subsection{Zero-Shot Visual Question Answering}
% Table \ref{tab:task_accuracy} presents the zero-shot model training performance of six plant disease understanding \gls{VQA} tasks. The random choice yields 24–30\% on the full benchmark, establishing a lower bound and expert performance as an upper bound. For general large GPT-4o and Gemini 2.5 Pro lead closed-source models with 90–92\% on HDC and 85–88\% on DI under the subset, and 78–92\% in general. Domain-specialized SCOLD outperforms all open-source \gls{VLMs}, scoring 97\% on DI and 84–89\% on CSI and SI. InternVL3.5 achieves 100\% on HDC (Tiny) but drops to 78\% on full benchmark, indicating confusion in prediction. General \gls{VLMs} (BLIP-2, CLIP, ALIGN, SigLIP2) achieve near-random accuracy on CSI, SI, and PC, highlighting the need for domain adaptation.

To evaluate LeafBench as a rigorous benchmark for disease visual understanding, we assessed a diverse set of thirteen vision–language models under a zero-shot protocol. The results, summarized in Table \ref{tab:task_accuracy} and Figure \ref{fig:test_overall}, confirm the benchmark’s challenges and its effectiveness in distinguishing between generic and domain-adapted reasoning.

\textbf{Dataset difficulty and discriminative power}: The LeafBench instruction set provides a significant challenge to off-the-shelf foundation models. As shown in Table \ref{tab:task_accuracy}, generic zero-shot baselines (e.g., CLIP, SigLIP2) frequently perform near the random chance lower bound ($\sim$24-30\%) on fine-grained tasks such as SI and PC. This ``generalization gap" confirms that LeafBench captures specialized pathological features that are underrepresented in general web-scale pre-training data. Conversely, the high performance of frontier proprietary models (GPT-4o, Gemini 2.5 Pro) and the domain-specialized SCOLD architecture (reaching 96.28\% on Disease Identification) demonstrates that the dataset is solvable and contains distinct, learnable signal—establishing a meaningful expert upper bound.

\textbf{Validation of task stratification}: The performance gradation across the six tasks confirms the hierarchical design of the benchmark. Models consistently achieve higher accuracy on the binary HDC task (reaching 90-100\%) compared to the complex taxonomic reasoning required for SNC, where even advanced models struggle to exceed 65\% accuracy. This validates that LeafBench successfully isolates distinct levels of diagnostic reasoning, from basic health screening to expert-level taxonomy.

\begin{landscape}
\begin{table*}[ht]
%\footnotesize
\centering
\caption{Accuracy (\%) of different models on six \gls{VQA} tasks in the plant pathology domain. Results are reported for both the \textit{Tiny} and \textit{All}, where \textit{Tiny} refers to the subset performance and \textit{All} refers to the complete benchmark. The best-performing \gls{LMMs} in each subset for plant disease domain \gls{LMMs} is \textbf{in-bold}, and the top-performing \gls{LMMs} is \underline{underline}.}

\label{tab:task_accuracy}
%\footnotesize
\begin{tabular}{|l|c|c|c|c|c|c|c|c|c|c|c|c|c|c|}
\toprule
\multirow{3}{*}{} & 
\multicolumn{2}{|c|}{\textbf{HDC}} & 
\multicolumn{2}{|c|}{\textbf{DC}} & 
\multicolumn{2}{|c|}{\textbf{CSI}} & 
\multicolumn{2}{|c|}{\textbf{SNC}} & 
\multicolumn{2}{|c|}{\textbf{PC}} & 
\multicolumn{2}{|c|}{\textbf{SI}} \\
\cmidrule(lr){2-3} \cmidrule(lr){4-5} \cmidrule(lr){6-7} 
\cmidrule(lr){8-9} \cmidrule(lr){10-11} \cmidrule(lr){12-13}
& \makecell{\textbf{Tiny}} & \makecell{\textbf{All}} 
& \makecell{\textbf{Tiny}} & \makecell{\textbf{All}} 
& \makecell{\textbf{Tiny}} & \makecell{\textbf{All}} 
& \makecell{\textbf{Tiny}} & \makecell{\textbf{All}} 
& \makecell{\textbf{Tiny}} & \makecell{\textbf{All}} 
& \makecell{\textbf{Tiny}} & \makecell{\textbf{All}} \\
\midrule
Random Choice & 50.00 & 50.47 & 27.91 & 24.07 & 30.00 & 26.20 & 22.46 & 25.51 & 22.22 & 26.14 & 28.53 & 26.18 \\
\midrule
GPT-4o \cite{gpt4o} & \underline{90.00} & \underline{92.48} & \underline{82.55} & \underline{85.27} & 87.27 & \textbf{85.58} & \textbf{66.46} & \textbf{65.27} & \textbf{44.44} & 56.47 & \underline{49.11} & \underline{51.64} \\ \hline
Gemini 2.5 Pro \cite{gemini} & 80.00 & 88.25 & 77.61 & 78.54 & \underline{88.18} & 83.21 & \underline{65.51} & \underline{64.89} & \textbf{44.44} & 51.23 & 48.26 & 48.99 \\ \hline
LLaVA 1.5 \cite{llava} & 50.00 & 80.23 & 40.70 & 33.29 & 50.00 & 49.93 & 24.31 & 29.11 & \textbf{44.44} & \underline{63.48} & 31.76 & 32.61 \\ \hline
LLaVA-NeXT \cite{llavanext} & 50.00 & 88.33 & 41.86 & 33.64 & 47.27 & 48.82 & 29.54 & 27.10 & \textbf{44.44} & \textbf{70.82} & 31.47 & 32.09 \\ \hline
Qwen 2.5 VL \cite{qwen25} & 80.00 & 81.05 & 52.33 & 52.60 & 63.64 & 63.06 & 41.85 & 41.50 & \textbf{44.44} & 60.92 & 40.88 & 43.14 \\ \hline
BLIP-2 \cite{blip-2} & 60.00 & 62.36 & 50.00 & 48.49 & 60.00 & 64.15 & 29.85 & 28.02 & \textbf{44.44} & 54.88 & 31.76 & 31.59 \\ \hline
CLIP \cite{clip} & 60.00 & 21.20 & 52.33 & 46.51 & 53.64 & 48.99 & 33.23 & 32.56 & 22.22 & 20.43 & 33.53 & 32.32 \\ \hline
BioCLIP \cite{bioclip} & 67.44 & 48.18 & 70.00 & 62.36 & 38.46 & 74.22 & 38.46 & 38.45 & \textbf{44.44} & 18.79 & 25.88 & 25.80 \\ \hline
SCOLD \cite{scold} & \underline{90.00} & \textbf{96.28} & \textbf{97.67} & \textbf{95.85} & \textbf{89.09} & \underline{84.73} & 38.46 & 41.64 & \underline{33.33} & 37.83 & \textbf{75.88} & \textbf{77.92} \\ \hline
SigLIP \cite{siglip} & 50.00 & 80.23 & 23.26 & 28.49 & 28.18 & 31.05 & 28.92 & 24.73 & 11.11 & 56.33 & 24.41 & 25.46 \\ \hline
SigLIP2 \cite{siglip2} & 50.00 & 27.75 & 34.88 & 26.67 & 23.64 & 28.80 & 24.31 & 27.20 & 22.22 & 16.67 & 20.00 & 22.75 \\ \hline
InternVL3.5 \cite{internvl3} & \textbf{100.00} & 78.37 & 45.35 & 35.54 & 43.64 & 32.52 & 35.69 & 31.45 & \underline{33.33} & 43.96 & 35.00 & 39.61 \\ \hline
ALIGN \cite{align} & 50.00 & 30.66 & 55.81 & 48.49 & 50.00 & 56.87 & 33.85 & 32.37 & 22.22 & 24.40 & 35.88 & 34.78 \\
\bottomrule
\end{tabular}
\end{table*}
\end{landscape}

\textbf{Reliability of the 'Tiny' subset}: Crucially, the evaluation confirms the statistical representativeness of our stratified 'Tiny' subset. The alignment in performance trends between the 'Tiny' (purple bars) and 'All' (blue bars) sets in Figure \ref{fig:test_overall} indicates that the subset accurately reflects the difficulty distribution of the full benchmark. This verifies the 'Tiny' set as a reliable, cost-effective proxy for evaluating computationally expensive commercial APIs or human expert performance without significant sampling bias.

\begin{figure}[ht]
    \centering
\includegraphics[width=\textwidth]{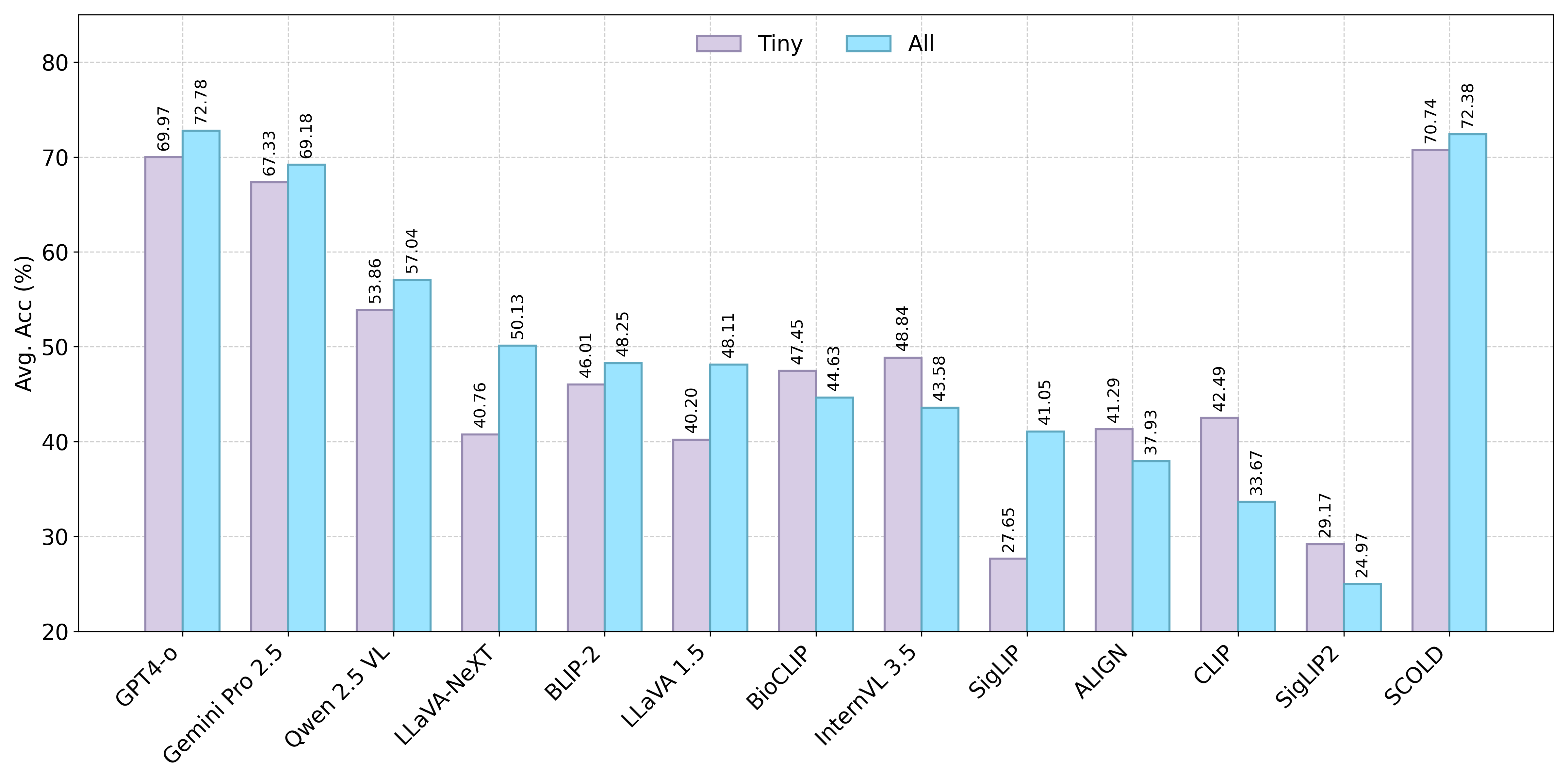}
    \caption{Average overall accuracy (\%) of different \gls{VLMs} on Tiny subset and full benchmark (All).}
    \label{fig:test_overall}
\end{figure}

% These findings underscore the critical role of domain specialization and model calibration in plant disease \gls{VQA}. As shown in Figure \ref{fig:test_overall}, the stark contrast between high-performing models (GPT-4o, Gemini 2.5 Pro, SCOLD) and general \gls{VLMs} (BLIP-2, CLIP, ALIGN, SigLIP2) is clearer. The contrast also highlights that large-scale pretraining on generic web data falls short for fine-grained plant diagnosis tasks. Instead, domain-specific pretraining on symptom-focused imagery—as demonstrated by SCOLD—or advanced zero-shot reasoning capabilities, as exemplified by GPT-4o, are crucial for achieving reliable performance across diverse diagnostic tasks.

\subsection{Limitations} \label{Limitations}
Despite its scale and multimodal richness, LeafNet experiences limitations in spatial and phenological coverage. Although the dataset was curated by aggregating images collected across 7 countries and 3 continents, broader geographic and environmental diversity is required to enhance the transferability of the proposed approach for application at a global scale. Moreover, most images are static captures in the visible spectrum, which may not fully capture the temporal dynamics of disease progression or the influence of diverse environmental stressors. While expert-curated textual annotations provide essential context for disease reasoning, they are largely confined to disease names, pathogen taxonomy, and brief symptom descriptions. This limited granularity can obscure variations in symptom severity, interactions among co-occurring stress factors, and latent phenotypic traits that emerge only under specific environmental conditions. As a result, models trained on LeafNet may struggle to generalize to novel field settings where phenotypic expressions evolve over time or extend beyond the visible spectral range.

\section{Conclusion} \label{Conclusion}
% In this paper, we present LeafNet and LeafBench, a rigorous foundation for advancing multimodal plant disease understanding by combining 186,000 expert-annotated leaf RGB images. The utility of the dataset and benchmark in six targeted \gls{VQA} tasks was evaluated with various models and parameter settings. Our results reveal that symptom‐focused pretraining, as SCOLD, rivals the zero‐shot reasoning of GPT-4o and outperforms generalist models on fine‐grained disease and symptom questions, underscoring the necessity of domain alignment. In vision task, end-to-end fine-tuning bridges the substantial gap between ImageNet features and plant pathology cues, with DenseNet121 achieving near‐perfect discrimination only after full adaptation. The divergent difficulty across tasks—where binary healthy–diseased decisions saturate at > 90\% accuracy while pathogen and species identification linger below 65\%—highlights the limits of current vision–language and vision-only models in resolving subtle visual taxonomies. Crucially, models that excel under extreme few-shot conditions do not always generalize to larger benchmarks, whereas those calibrated for both training regimes deliver consistent performance. By demonstrating these insights at scale, this work illuminates the intertwined roles of multimodal annotations, domain-specialized pretraining, and training strategy in deploying novel \gls{AI} for agriculture—and points toward more resilient, explainable systems that can adapt to the nuanced demands of field diagnostics.

In this paper, we introduce paired LeafNet and LeafBench as a foundational dataset designed to advance multimodal understanding of plant diseases. The dataset comprises over 186,000 expertly annotated image–metadata pairs. Through comprehensive benchmarking, we demonstrate both the high quality of the data and its critical role in revealing the limitations of current generalist AI systems in precision plant disease detection and management.

The pronounced performance gap between linear probing and full fine-tuning provides strong evidence that LeafNet represents a distinct visual–semantic distribution not captured by broadly pre-trained models. While consistently high convergence during fine-tuning across diverse architectures confirms the coherence and richness of our expert annotations, the failure of off-the-shelf ImageNet features highlights a fundamental mismatch: plant pathology demands specialized feature representations beyond those learned for general object recognition. These results establish LeafNet not merely as an image–text collection, but as a necessary substrate for training domain-adaptive visual backbones. Furthermore, the stratified design of LeafBench—spanning binary screening tasks to fine-grained taxonomic reasoning—systematically delineates the boundaries of current machine perception. Although modern models achieve strong performance in healthy–diseased classification, they struggle substantially with the nuanced visual reasoning required for disease and symptom identification. This controlled escalation in task complexity demonstrates that LeafBench is a rigorous benchmark for assessing true diagnostic understanding rather than superficial pattern matching.

Collectively, this work reframes the progress in plant disease diagnostics by shifting the focus from model architecture to data-centric AI. Our findings suggest that advances in disease diagnostics will not arise solely from scaling generalist models, but from learning systems grounded in high-quality, domain-aligned data. By providing this rigorous testbed, we aim to accelerate the development of robust and interpretable AI systems capable of addressing the nuanced demands of plant disease management, crop protection, and global food security.

\section{Future work} \label{future}
In the future, expanding the dataset and enriching the annotation schema presents a critical avenue for advancing leaf disease understanding. Although the current dataset was collected from different countries and environmental conditions, a larger and more geographically and environmentally diverse dataset will enhance the accuracy, reliability, and transferability of the method in real-world applications. Furthermore, expanding text annotations to encompass structured ontologies, such as crop, disease severity scales, and growth stage metadata, will enable instruction-tuned multimodal models to interpret complex agronomic queries. Constructing a large instruction-tuning dataset, wherein each example pairs an image with detailed, natural-language explanations and step-by-step diagnostic reasoning, can catalyze the development of the next generation of leaf-focused multimodal foundation models. Such a dataset would enable models to not only recognize visual cues but also articulate biological mechanisms and management recommendations in a manner analogous to expert agronomists.

Another promising direction involves integrating multi-temporal and multispectral imagery into LeafBench. Time-lapse sequences capturing disease onset, progression, and recovery would enable models to learn temporal dynamics and forecast disease trajectories. In addition, incorporating near-infrared, thermal, and hyperspectral bands can reveal early stress signatures that are imperceptible in broad-band RGB images. By unifying spatial, spectral, and temporal modalities, future iterations of LeafNet can facilitate the development of more robust, real-time disease monitoring systems that detect outbreaks in their earlier stages and guide precision interventions, plant protection strategies, and effective disease management. This multi-dimensional extension will be essential for deploying \gls{AI} in dynamic, resource-constrained agricultural environments.

\section*{CRediT authorship contribution statement}
\textbf{Khang Nguyen Quoc} contributed to conceptualization, data curation, methodology, software, model development, data analysis and visualization, and writing – original draft. \textbf{Phuong D. Dao} contributed to methodology, supervision, validation, and writing – review \& editing. \textbf{Luyl-Da Quach} contributed to methodology, project administration, supervision, validation, and writing – review \& editing.

\section*{Declaration of Competing Interest}
The authors declare no known competing conflicts of interest that could have influenced the work reported in this paper.

\section*{Acknowledgment}
This research was supported by the Hyundai Motor Chung Mong-Koo Foundation Global Scholarship (GSS-25-02120).

\section*{Data Availability}

LeafNet and LeafBench are publicly available at \url{https://huggingface.co/collections/enalis/leafsight}

%\bibliographystyle{ieeetr}
%\bibliography{cas-refs}

\appendix
\setcounter{figure}{0}
\setcounter{table}{0}
\section{Supplementary Materials}
\label{sec:supp}

\begin{figure}[H]
    \centering
\includegraphics[width=0.9\textwidth]{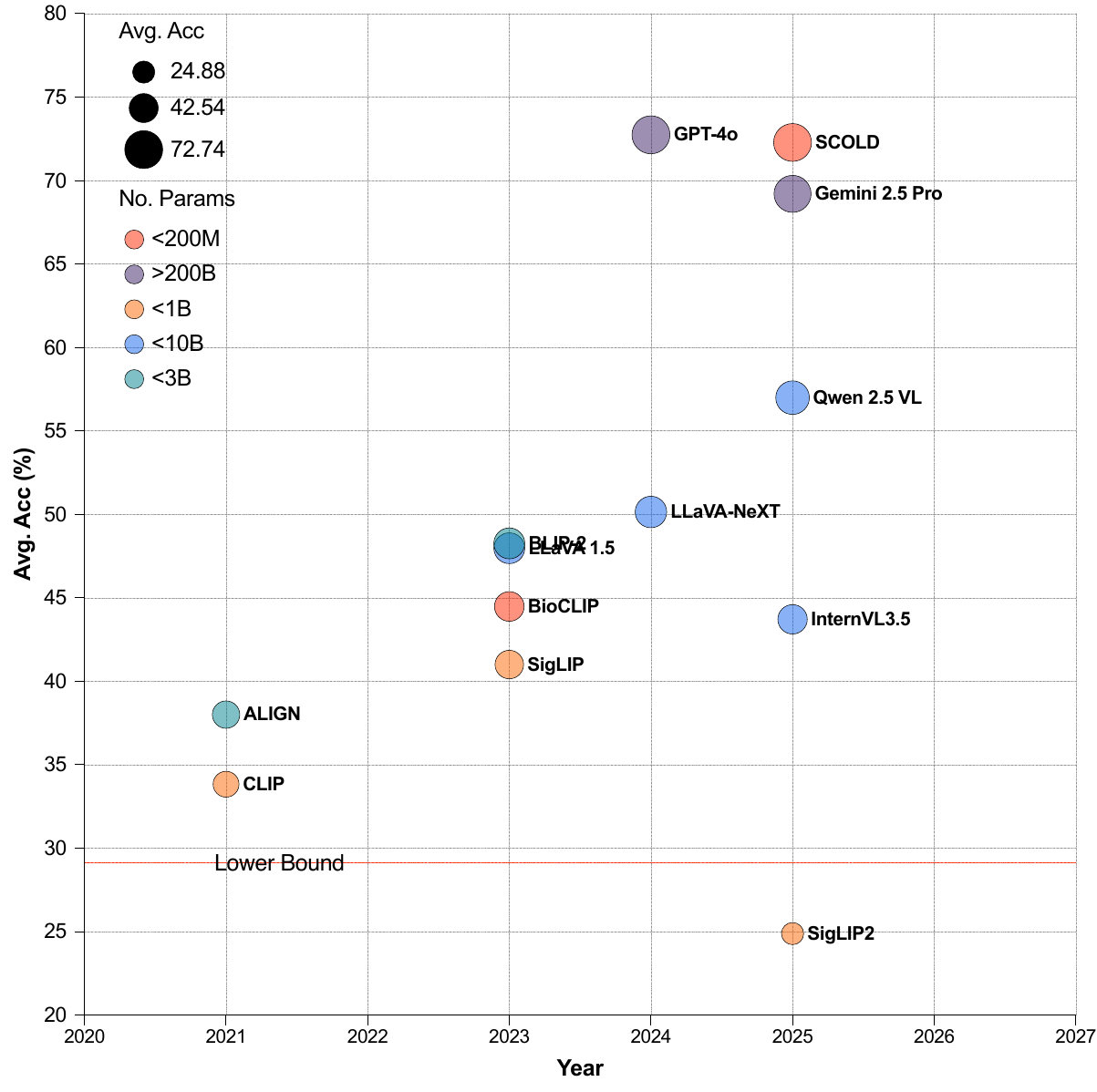}
    \caption{The performance of \gls{VLMs} over time on the LeafBench benchmark. The random choice baseline (\textcolor{red}{red}) serves as the lower bound.
}
    \label{fig:vlm_all_avg}
\end{figure}

\begin{figure}[H]
    \centering
    \includegraphics[width=0.8\linewidth]{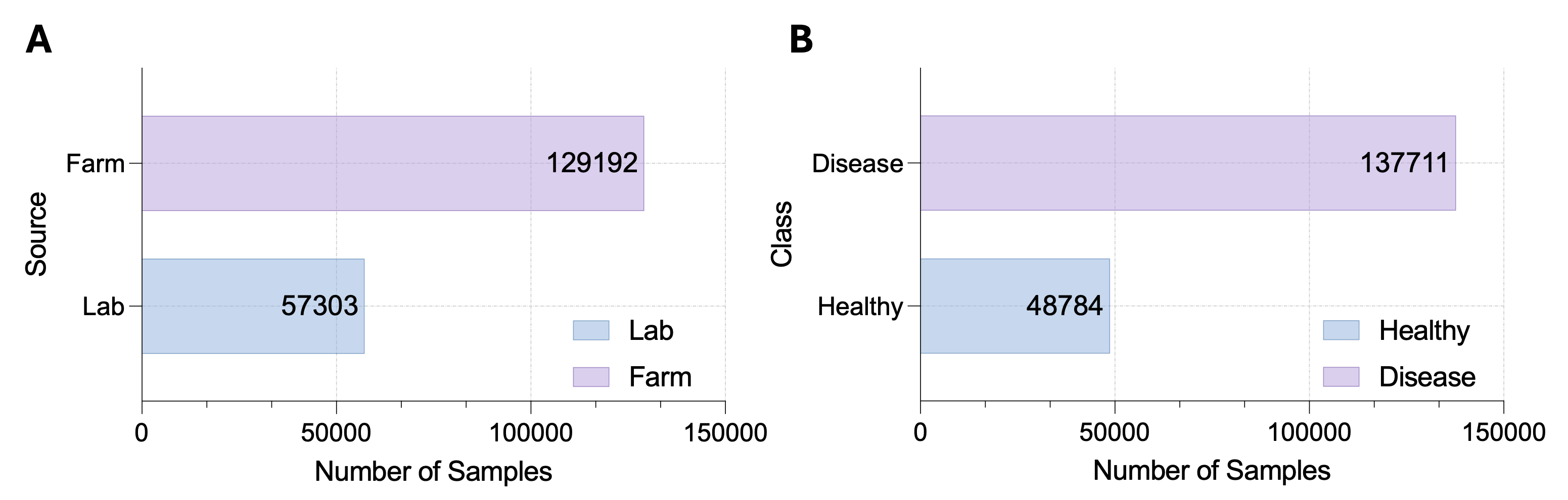}
    \caption{Distribution of image acquisition sources (laboratory vs. farm).}
    \label{fig:dh_farm_lab}
\end{figure}

\begin{figure}[H]
    \centering
    \begin{subfigure}[t]{0.45\textwidth}
        \centering
        \includegraphics[width=\textwidth]{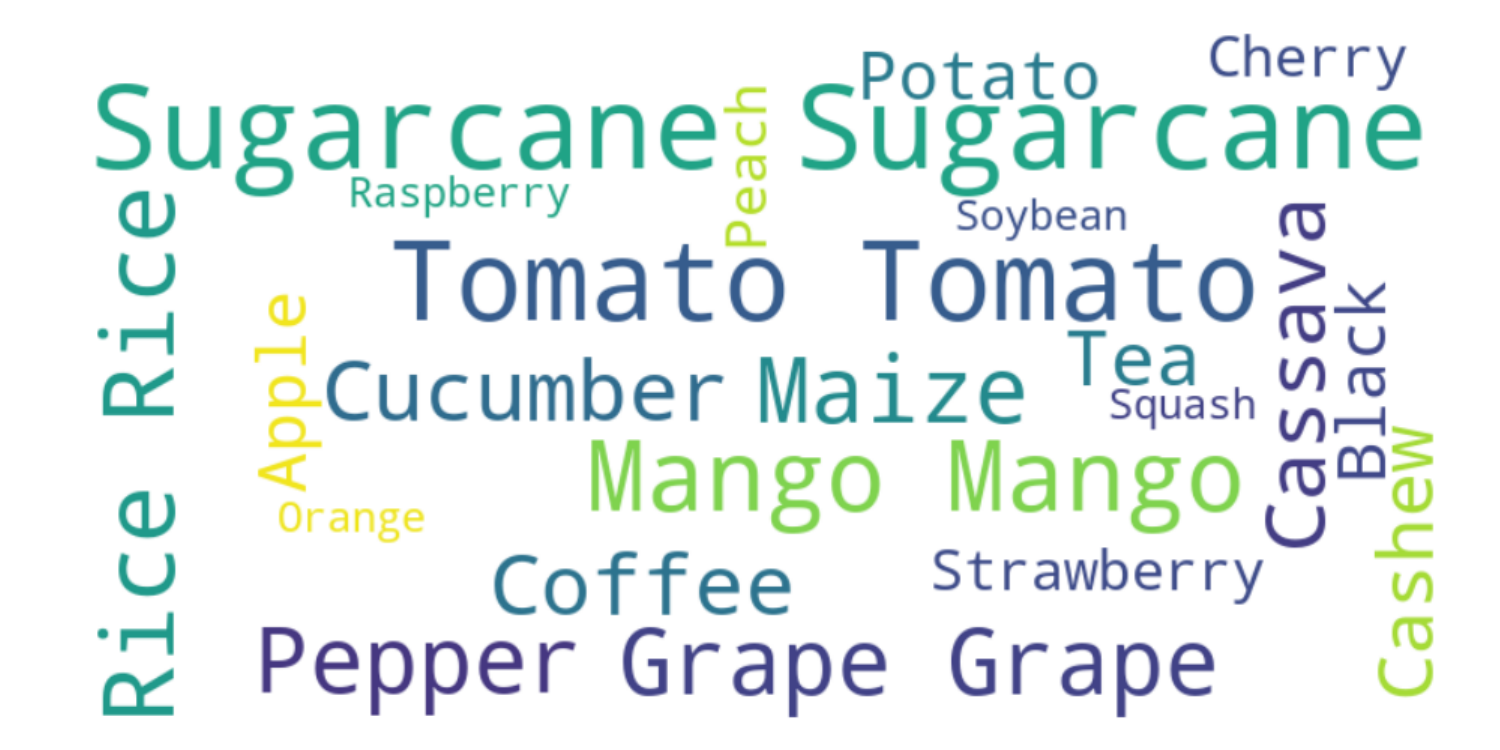}
        \caption{Crop}
    \end{subfigure}
    \hfill
    \begin{subfigure}[t]{0.45\textwidth}
        \centering
        \includegraphics[width=\textwidth]{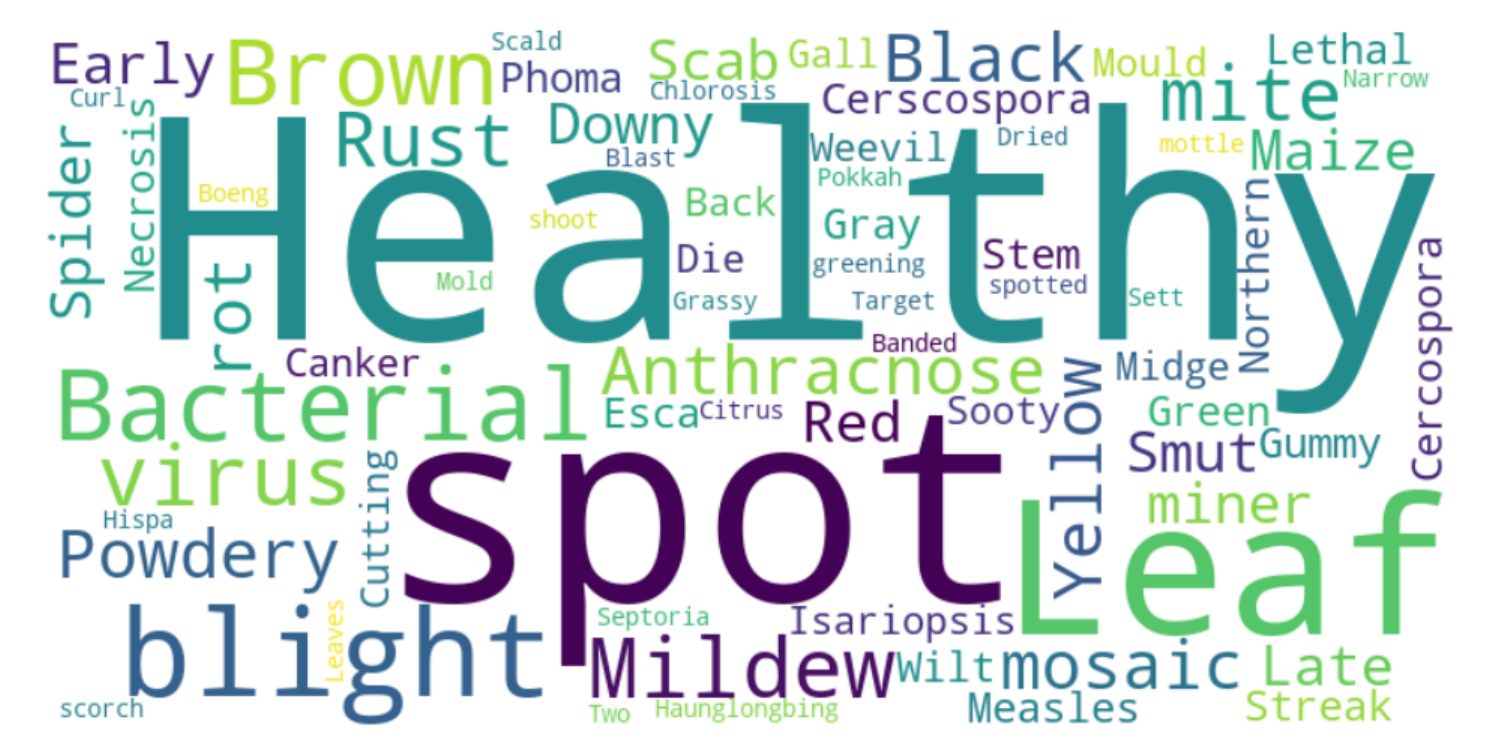}
        \caption{Disease}
    \end{subfigure}
    
    \vspace{1em} 
    \begin{subfigure}[t]{0.45\textwidth}
        \centering
        \includegraphics[width=\textwidth]{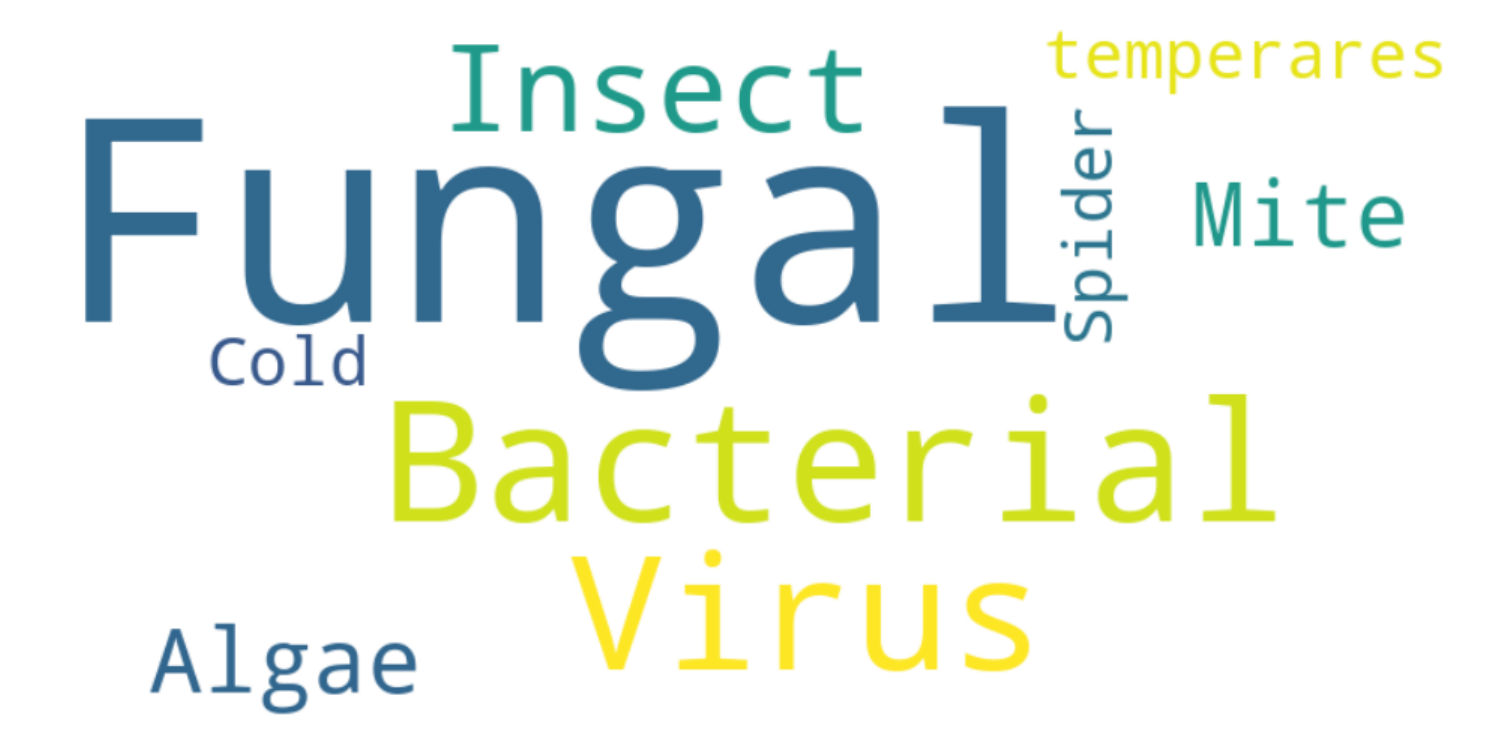}
        \caption{Pathogen}
    \end{subfigure}
    \hfill
    \begin{subfigure}[t]{0.45\textwidth}
        \centering
        \includegraphics[width=\textwidth]{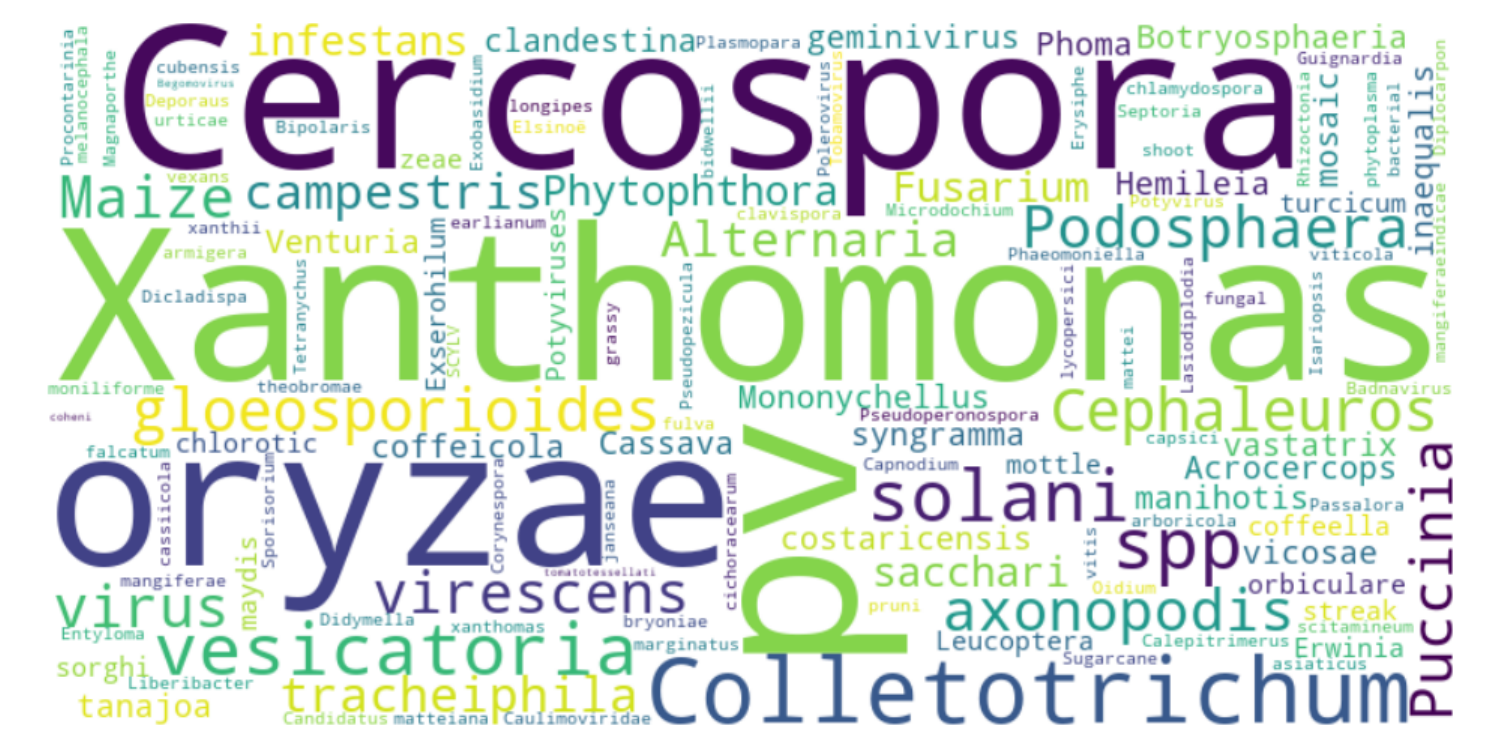}
        \caption{Scientific Name}
    \end{subfigure}

    \vspace{1em} 
    \begin{subfigure}[t]{0.45\textwidth}
        \centering
        \includegraphics[width=\textwidth]{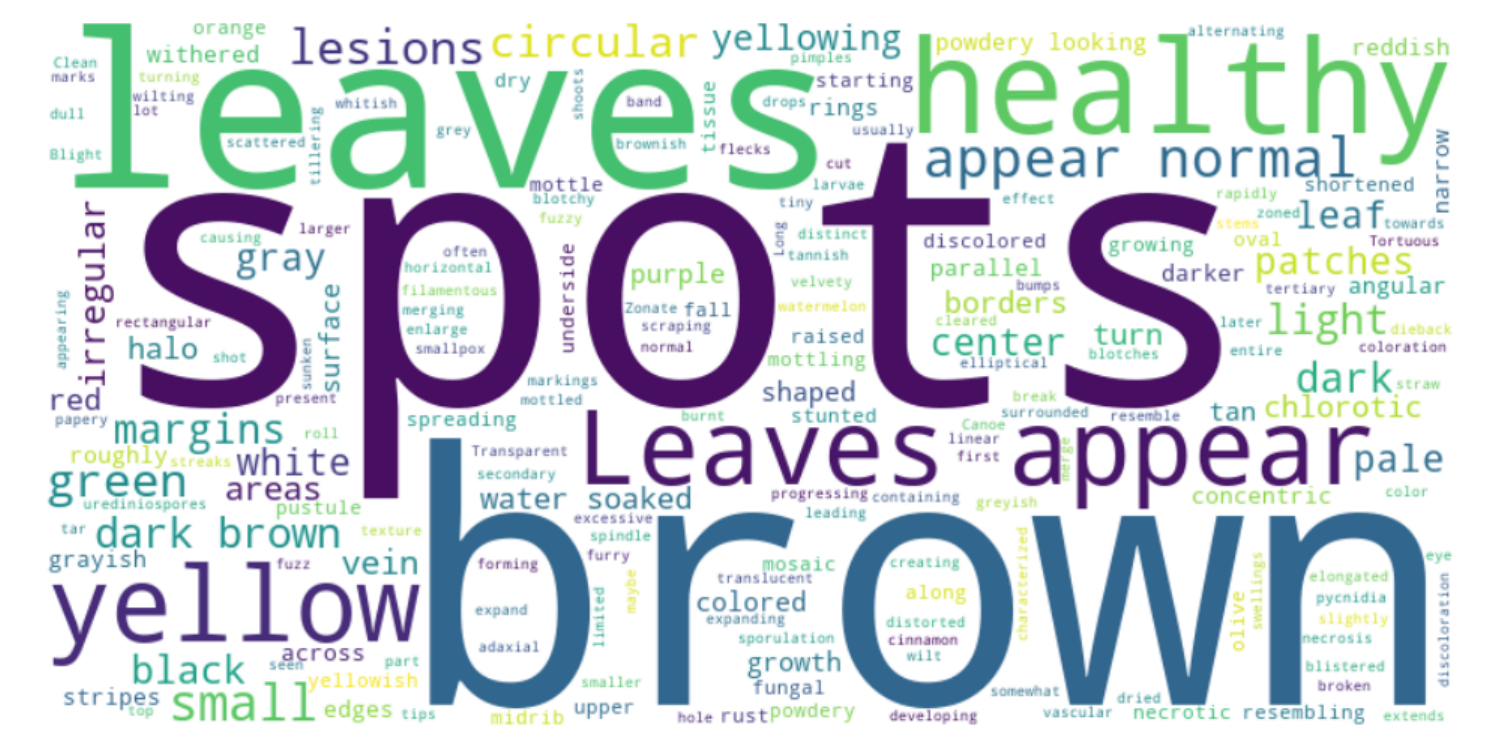}
        \caption{Symptom Description}
    \end{subfigure}
    \hfill
    \begin{subfigure}[t]{0.45\textwidth}
        \centering
        \includegraphics[width=\textwidth]{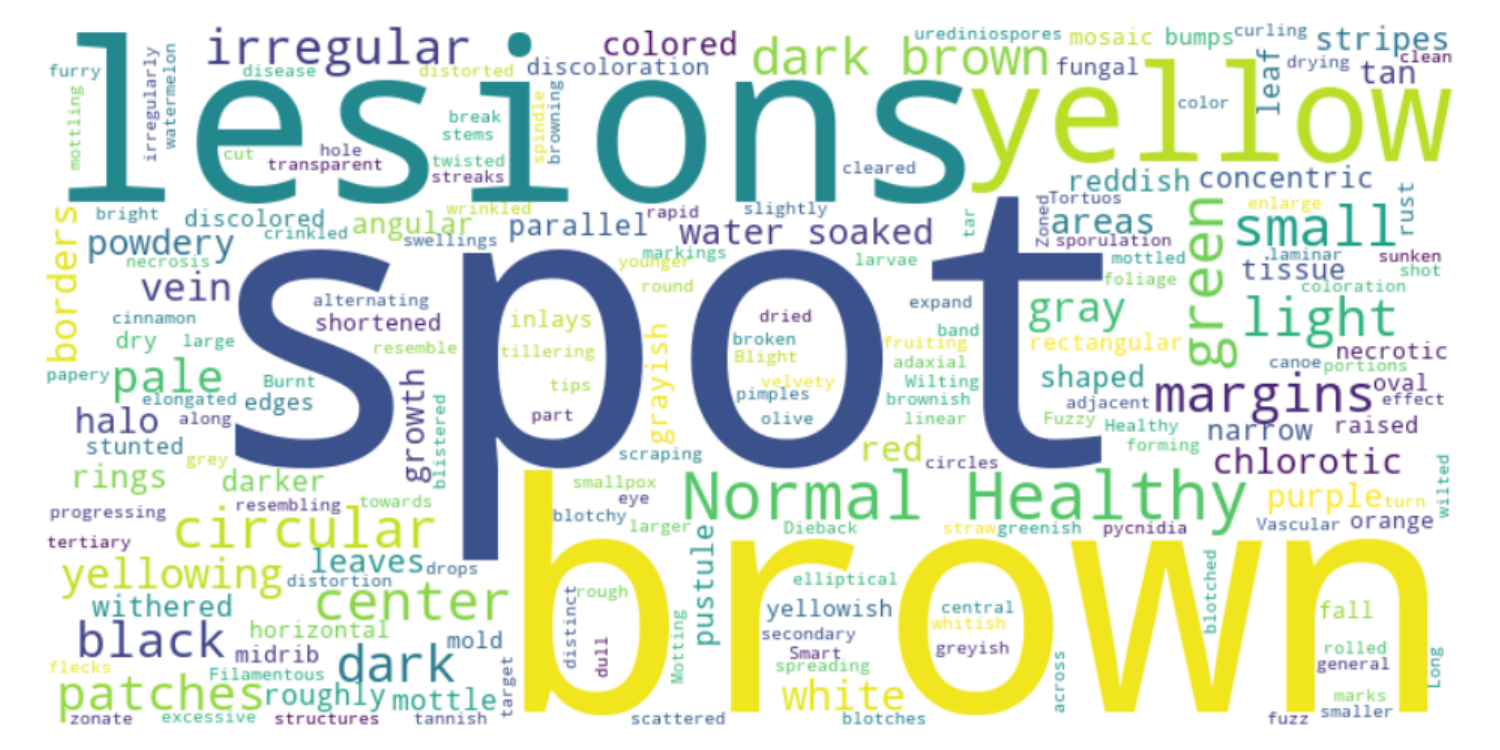}
        \caption{Main Symptom}
    \end{subfigure}
    \caption{Word clouds representing the lexical diversity in LeafNet across five facets: crop, disease, pathogen, scientific name, and symptom descriptions.}
    \label{fig:LeafNet_wordclouds}
\end{figure}

\begin{figure}[!htbp]
  \centering
  \begin{subfigure}{0.32\textwidth}
    \centering
    \includegraphics[width=\linewidth]{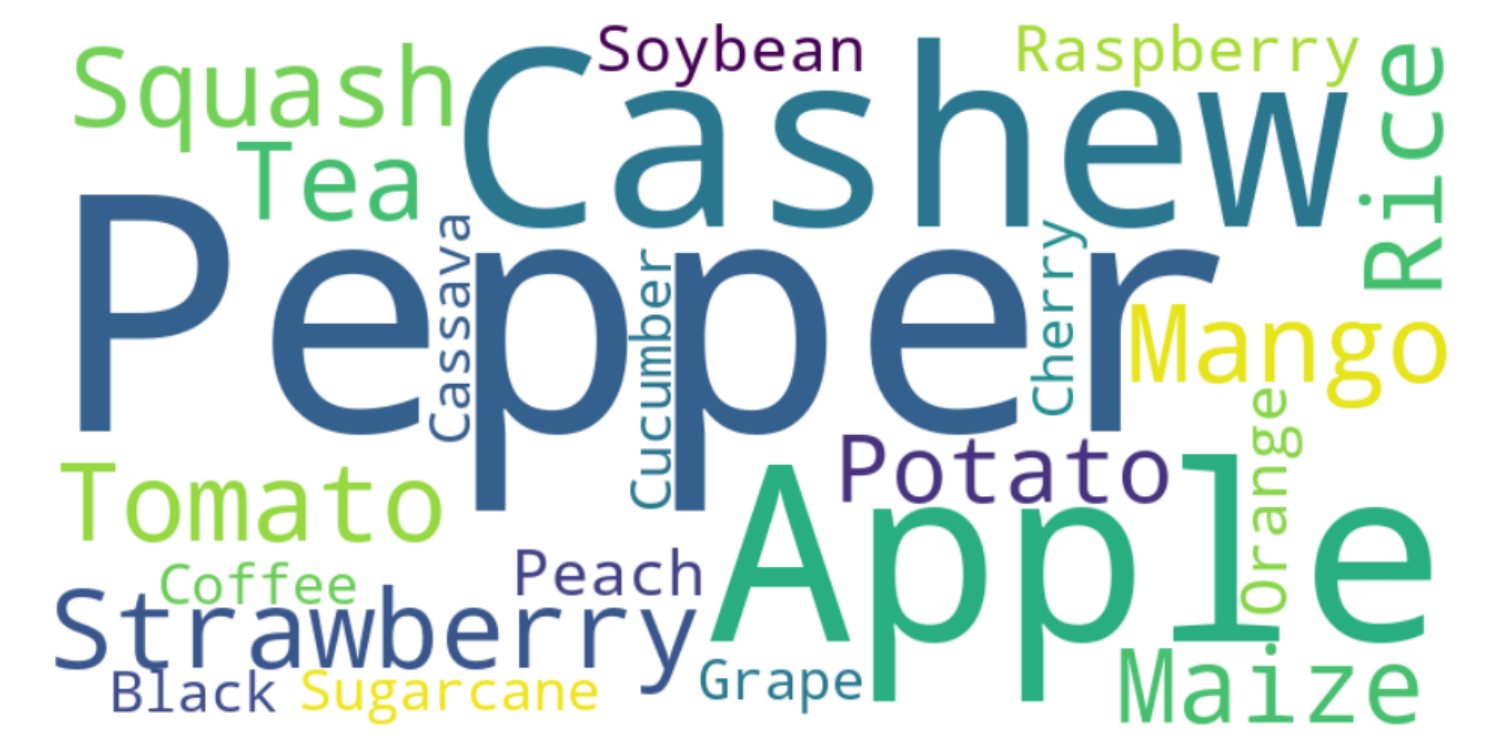}
    \caption{CSI}
  \end{subfigure}\hfill
  \begin{subfigure}{0.32\textwidth}
    \centering
    \includegraphics[width=\linewidth]{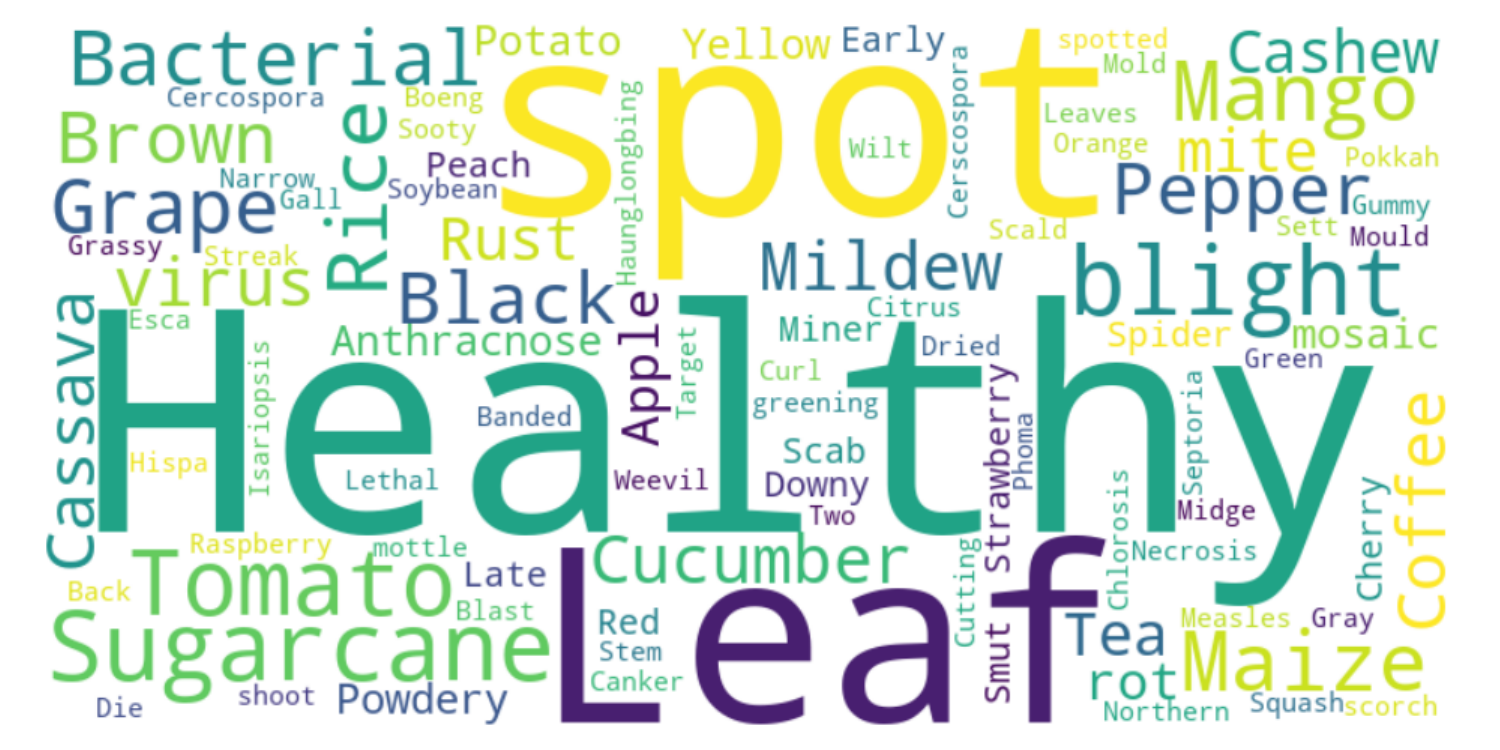}
    \caption{DI}
  \end{subfigure}\hfill
  \begin{subfigure}{0.32\textwidth}
    \centering
    \includegraphics[width=\linewidth]{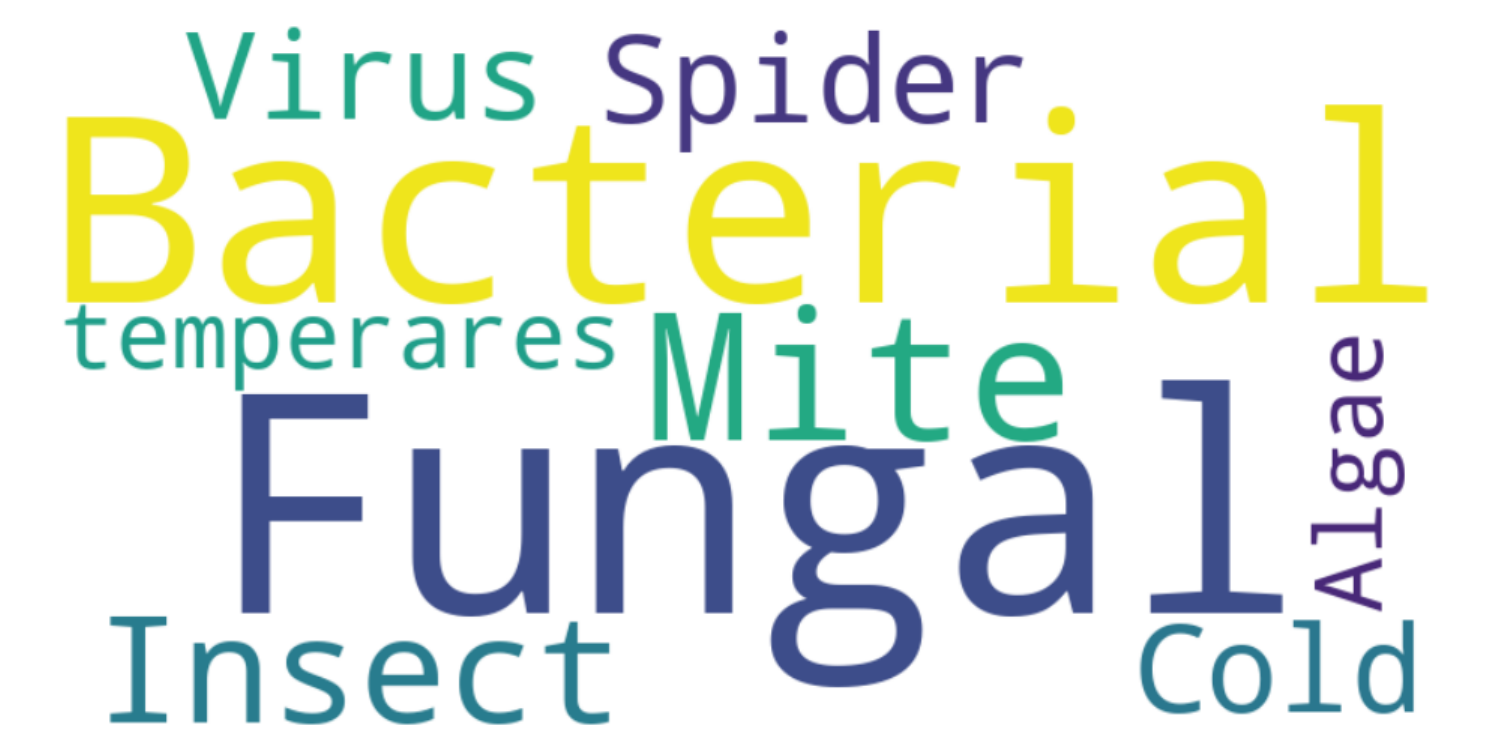}
    \caption{PC}
  \end{subfigure}
  \vspace{4mm} 
  \begin{subfigure}{0.45\textwidth}
    \centering
    \includegraphics[width=\linewidth]{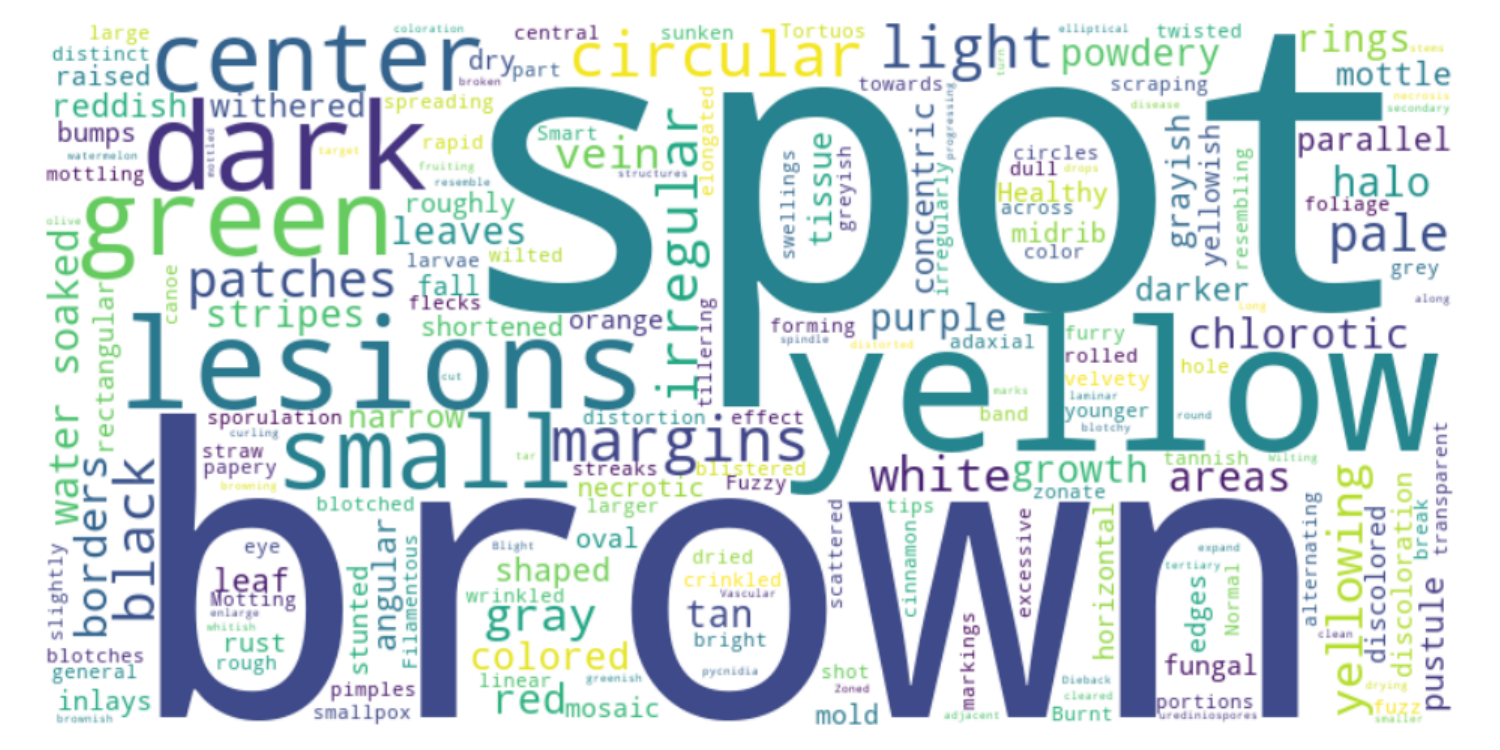}
    \caption{SI}
  \end{subfigure}\hfill
  \begin{subfigure}{0.45\textwidth}
    \centering
    \includegraphics[width=\linewidth]{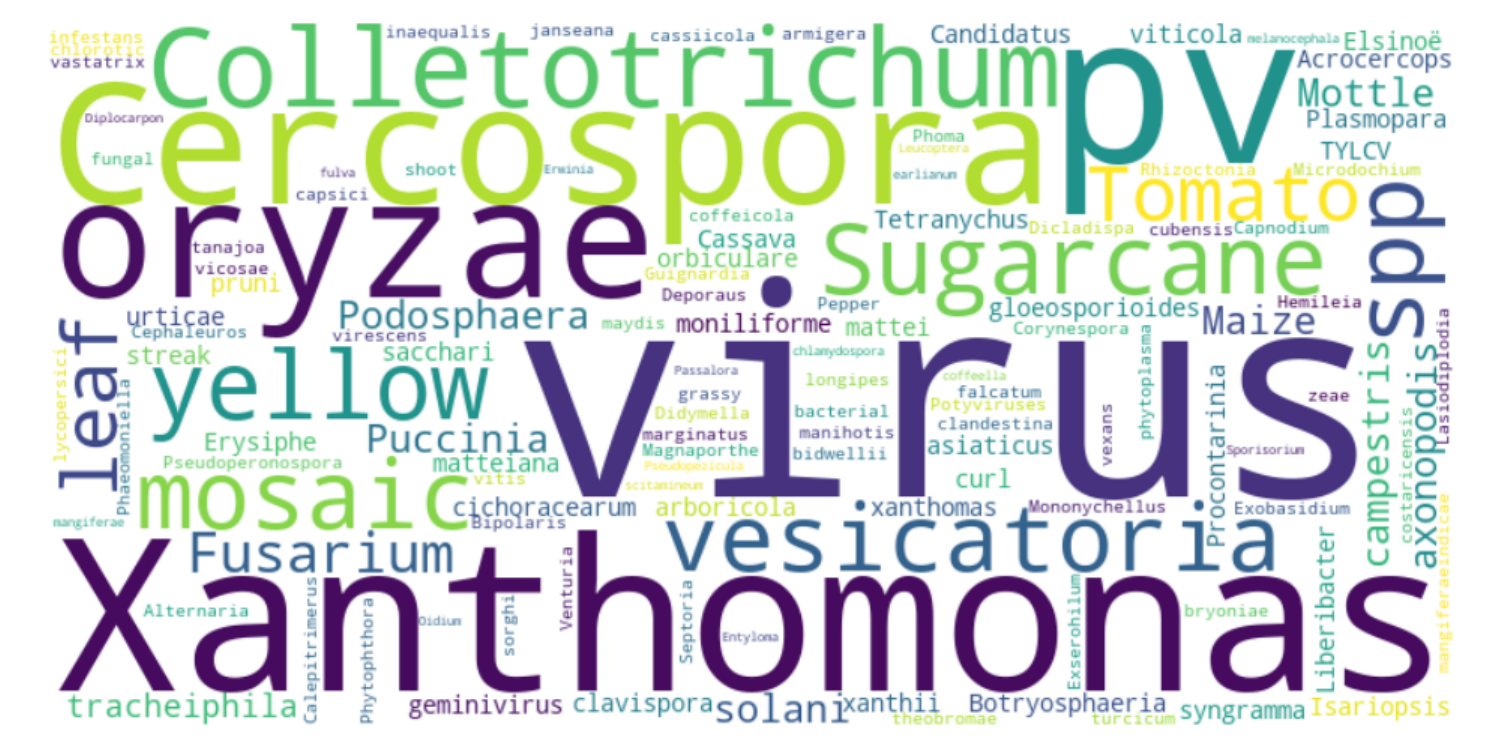}
    \caption{SNC}
  \end{subfigure}
  \caption{Word clouds representing the answer options available in the LeafBench benchmark.}
  \label{fig:answer_dist}
\end{figure}

\begin{table}[H]
\centering
\caption{Hyperparameter setup for image classification.}
\label{tab:hyperparameters}
\begin{tabular}{|c|c|}
\hline
\textbf{Hyperparameter} & \textbf{Value} \\ \hline
Batch Size              & 32 \\ \hline
Learning Rate           & $1 \times 10^{-3}$ \\ \hline
Optimizer               & AdamW \\ \hline
Epochs                  & 100 \\ \hline
Weight Decay            & 0.05 \\ \hline
Scheduler               & Cosine Annealing \\ \hline
Input Resolution        & $224 \times 224$ \\ \hline

\end{tabular}
\end{table}

\end{document}